%% file: main.tex
\documentclass{article}

\usepackage[utf8]{inputenc} 
\usepackage[T1]{fontenc}    
\usepackage{hyperref}       
\usepackage{url}            
\usepackage{booktabs}       
\usepackage{amsfonts}       
\usepackage{nicefrac}       
\usepackage{microtype}      
\usepackage{xcolor}         
\usepackage{graphicx}
\usepackage{subcaption}
\usepackage{caption}
\usepackage{hyphenat}
\usepackage{setspace}
\usepackage{amsmath}
\usepackage{amssymb}
\usepackage{mathtools}
\usepackage{amsthm}
\usepackage{longtable}
\usepackage{multirow}
\usepackage{float}
\usepackage{tabularx}
\usepackage{enumitem}
\usepackage{threeparttable}
\usepackage{algorithm}
\usepackage{algorithmic}
\usepackage{titletoc}
\usepackage{wrapfig}
\usepackage{makecell}
\usepackage[table]{xcolor}
\usepackage[most]{tcolorbox}

\newtcolorbox{takeaway}[1][]{
  colback=blue!5, colframe=black, boxrule=0.7pt, arc=2mm,
  left=1.5mm, right=1.5mm, top=1mm, bottom=1mm,
  fonttitle=\small\bfseries, title=Takeaway, #1
}

\usepackage[preprint]{my_style}

\usepackage{tcolorbox}
\tcbuselibrary{skins, breakable, listings}

\definecolor{brandblue}{RGB}{57,95,207}
\definecolor{linkblue}{HTML}{0064E0}
\definecolor{textgray}{HTML}{1C2B33}
\definecolor{boxbg}{HTML}{F1F4F7}

\hypersetup{
  colorlinks=true,
  linkcolor=brandblue,
  citecolor=brandblue,
  urlcolor=linkblue
}

\newcommand{\paperTitle}{Rethinking Relative Policy Differences Transfer of On-Policy Distillation}

\newcommand{\paperAuthors}{%
  {\sffamily\bfseries Rong Wu$^{*,1,2}$}, 
  {\sffamily\bfseries Daocheng Fu$^{*,2,3}$}, 
  {\sffamily\bfseries Yu Yang$^{1,2}$},
  {\sffamily\bfseries Xuemeng Yang$^{2}$},
  {\sffamily\bfseries Siqi Li$^{1,2}$},
  {\sffamily\bfseries Shu Zou$^{4}$},
  {\sffamily\bfseries Yao Zhu$^{5}$},
  {\sffamily\bfseries Botian Shi$^{2}$},
  {\sffamily\bfseries Bo Zhang$^{2}$},
  {\sffamily\bfseries Cong Zhang$^{1}$},
  {\sffamily\bfseries Pinlong Cai$^{\dag,2}$},
  {\sffamily\bfseries Jianbiao Mei$^{\dag,2}$},
}

\newcommand{\paperAffiliations}{%
  {\normalsize $^1$ Zhejiang University}, 
  {\normalsize $^2$ Shanghai Artificial Intelligence Laboratory}
  {\normalsize $^3$ Fudan University}, 
  {\normalsize $^4$ The Australian National University}
  {\normalsize $^5$ Renmin University of China}
}

\newcommand{\paperNotes}{%
  {\small $^*$ Equal Contribution}, {\small $^\dag$ Corresponding Author}%
}

\newcommand{\githubLink}{\url{https://github.com/KnowledgeXLab/A2-OPD}}
\newcommand{\publishDate}{\today}

\newcommand{\renderFrontBox}{%
    \tcbset{
    enhanced, frame hidden,
    colback=boxbg,
    left=0.5cm, right=0.5cm, top=0.5cm, bottom=0.5cm,
    arc=16pt,
    before skip=0pt,
    grow to left by=1.5pt, grow to right by=1.5pt,
    overlay={
    \node[anchor=north east, at=(frame.north east), xshift=-2.3cm, yshift=-0.5cm] 
        {\includegraphics[height=1cm]{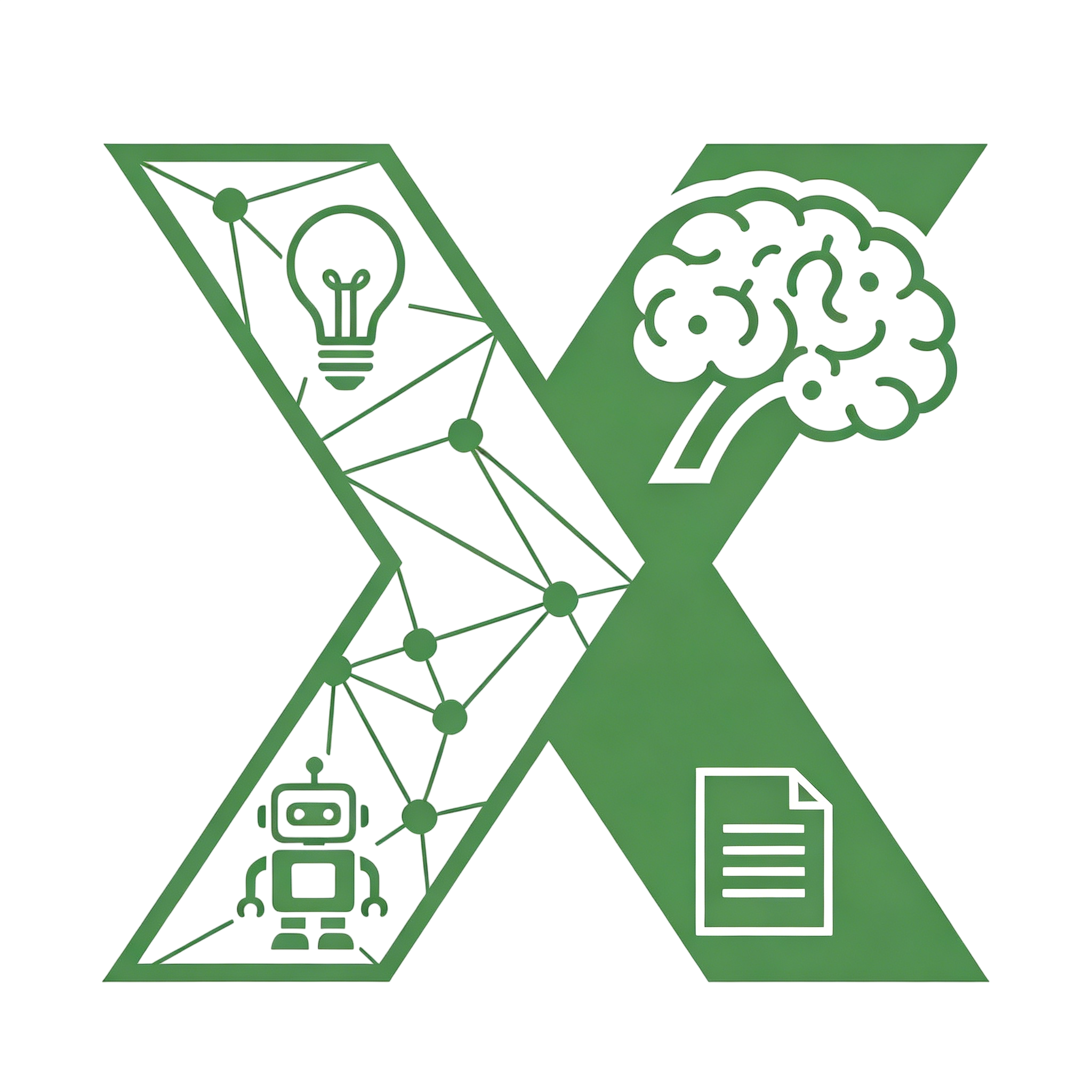}};
    \node[anchor=north east, at=(frame.north east), xshift=-0.5cm, yshift=-0.5cm] 
        {\includegraphics[height=1cm]{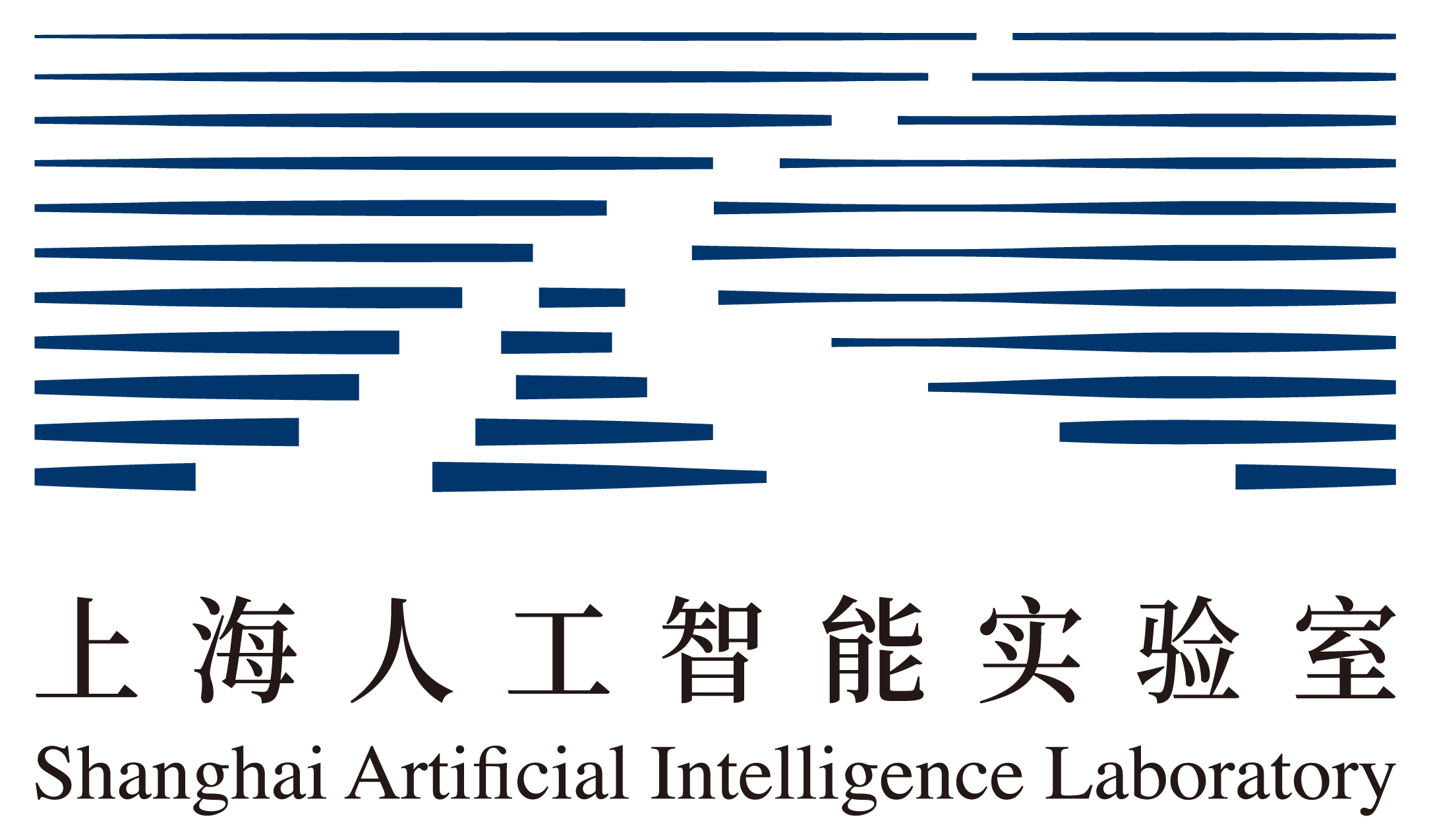}};
    }
  }%
  \begin{tcolorbox}
    \setlength{\parindent}{0cm}
    \setlength{\parskip}{0.5cm}
    {
      \setlength{\parskip}{0cm}
      \raggedright
      \nohyphens
      {
        \vskip 1cm 
        \setstretch{1.4}
        {\huge\sffamily\bfseries\textcolor{black}{\paperTitle}}\par
      }
      \vskip 0.25cm
      \paperAuthors\par
      \vskip 0.35cm
      \paperAffiliations\par
      \vskip 0.08cm
      \paperNotes\par
    }
    \vskip 0.2cm
    {\color{textgray}%

\input{sections/abstract}
\par}
    \vskip 0.4cm
    {
      \setlength{\parskip}{0cm}
      {\small {\sffamily\bfseries \raisebox{-0.2em}{\includegraphics[width=0.025\linewidth]{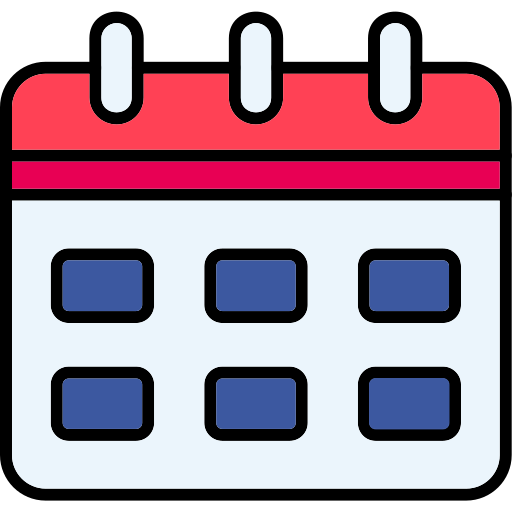}}~~Date:} \publishDate}\par%
      \vskip 0.08cm%
      {\small {\sffamily\bfseries \raisebox{-0.2em}{\includegraphics[width=0.025\linewidth]{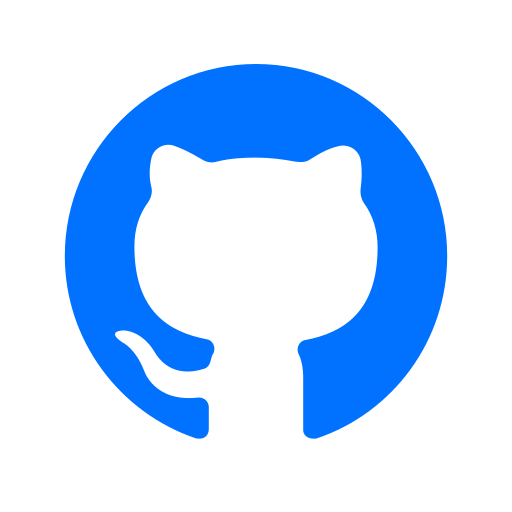}}~~Github Repo:} \githubLink}
    }
  \end{tcolorbox}
  \tcbset{reset}
}

\begin{document}

\newgeometry{top=1in, bottom=0.75in, textwidth=6.3in, textheight=9in}
\renderFrontBox



\input{sections/introduction}
\input{sections/related_works}
\input{sections/method}
\input{sections/experiment}

\input{sections/conclusion}

\clearpage
{
\bibliographystyle{unsrt}  
\bibliography{preprint}
}

\clearpage
\newgeometry{
  textheight=9in, textwidth=5.5in, top=1in,
  headheight=12pt, headsep=25pt, footskip=30pt
}

\input{sections/appendix}

\end{document}

%% file: sections/abstract.tex
\begin{abstract}
On-policy distillation (OPD) trains a target policy on its own rollouts using token-level supervision from a stronger teacher, but typically relies on the teacher being substantially more capable than the target. Recent work relaxes this requirement by transferring the relative policy difference between an expert and a reference policy rather than the expert's absolute distribution. Yet it remains unclear which differences provide useful supervision and which targets can benefit from them. We develop a unified view in which the log-probability ratio between a proxy expert and a proxy reference defines an implicit reward for KL-regularized policy optimization. This formulation induces an on-policy update for the target, recovers conventional OPD as a special case, and enables any-to-any transfer across models that differ in scale, training stage or generation. Our systematic experiments yield three findings. First, a stronger expert is not necessarily a better source of supervision: differences induced by task-specific training, including reinforcement learning and supervised fine-tuning, transfer consistently, whereas those arising from model scale alone transfer weakly or not at all. Second, the expert and reference serve complementary roles: the expert provides the capability to be transferred, while the reference filters out shared behavior and isolates the relevant residual. Third, even a useful difference does not transfer to every target: success requires the induced update to move the target toward the expert and generalize beyond sampled rollouts. Raw policy distance does not predict transferability, whereas an appropriate reference can rescue otherwise unsuccessful transfer. These results characterize when relative policy differences provide effective supervision for OPD.
\end{abstract}

%% file: sections/introduction.tex
\section{Introduction}
On-policy distillation (OPD) has emerged as an important approach to post-training large language models (LLM)~\cite{agarwal2024policy,lu2025opd,yang2025qwen3,deepseekai2026deepseekv4highlyefficientmilliontoken}. By eliciting teacher guidance on samples generated by the student itself, OPD mitigates the distribution shift between offline data and the student policy while providing more targeted training signals. However, conventional OPD directly encourages the student policy to approximate the teacher policy, making its effectiveness dependent on the teacher’s performance advantage over the student. When the two models have comparable capabilities, or when the teacher is weaker, distillation often yields limited gains and may even impair the student’s existing capabilities. Consequently, OPD typically requires a substantially stronger teacher~\cite{burns2023weaktostronggeneralizationelicitingstrong, yuan2026incentivizingstrongreasoningweak}, restricting its applicability and limiting the reuse of existing post-training outcomes.

To reduce this dependence on the teacher’s absolute capability, recent studies have explored \textbf{relative policy update transfer}~\cite{heo2026policy, feng2026weak, fu2026proxyopd}. Rather than directly aligning the output distributions of student and teacher, this approach extracts the relative policy update induced by post-training by comparing the teacher model before and after training, and then transfers this update to the proxy target model. It therefore does not require the teacher to substantially outperform the student; it only requires the post-trained teacher to outperform its corresponding base model. By shifting the transfer target from the teacher's output distribution to the policy change induced by its post-training, this paradigm broadens the applicability of OPD across models of different scales and architectures, and offers a promising way to reuse costly post-training outcomes.

Despite its promise, the scope of relative policy update transfer remains poorly
understood. Existing studies typically consider a narrow class of source
pairs, in which the expert is obtained by applying reinforcement learning (RL)~\cite{guo2025deepseekr1} to
the reference model. It is therefore unclear whether policy differences
arising from other sources, such as supervised fine-tuning (SFT)~\cite{ouyang2022training,pmlr-v267-chu25c} or model scaling~\cite{kaplan2020scalinglawsneurallanguage,Hoffmann2022training},
provide equally useful supervision. Moreover, prior work has largely focused
on demonstrating transfer gains, with limited analysis of how the expert,
reference and target jointly determine the outcome. As a result, the
conditions under which a policy difference is useful and transferable remain
unclear.

In this work, we revisit relative policy difference transfer through the lens of implicit reward and formally derive its transfer objective. The derivation interprets the policy difference between a proxy expert and a proxy reference model as the supervision transferred to a target model, while clarifying the distinct roles of all three models. It further separates two questions that are often
conflated: whether a source pair encodes a useful improvement~\cite{NEURIPS2025_537d5aa7}, and whether a
given target can realize that improvement. Guided by this derivation, we
conduct a systematic empirical study across policy differences induced by
RL, SFT and model scaling, as well as
across target models with different initial policies. Specifically, we ask:
\textit{(i) Which policy differences provide useful supervision?
(ii) Which capabilities of the expert are transferred, and what role does the
reference play? (iii) Which target models can effectively realize a useful
policy difference?}

Our results reveal three main findings. \textbf{(i)} a stronger model is not
necessarily a better source of supervision. Improvements acquired through
task-directed training transfer consistently, whether induced by RL or SFT, whereas differences caused
by model scaling alone transfer weakly or unreliably. \textbf{(ii)} the proxy expert and proxy reference play distinct and complementary roles: the expert provides the
capability, while the reference filters out behaviors already shared by the two models, exposing the remaining improvement for transfer~\cite{li-etal-2023-contrastive}. \textbf{(iii)} a useful policy difference is not universally transferable. Transfer succeeds when the induced update moves the target toward the expert and generalizes beyond the training samples. Raw policy distance does not predict this compatibility; critically, an appropriate reference can recover an otherwise failed transfer.

Our contributions are:
\begin{itemize}[leftmargin=1.2em, itemsep=1pt, topsep=2pt]
    \item We formally derive relative policy difference transfer from implicit rewards
    and formulate an any-to-any OPD objective that distinguishes the roles of
    the proxy expert, reference, and target.

    \item We systematically evaluate policy differences induced by RL, SFT,
    and model scaling across diverse targets, showing how the expert,
    reference, and target jointly shape transfer outcomes.

    \item We characterize when policy differences are useful and transferable,
    identifying task-directed training, reference filtering, and target
    compatibility as the key factors governing transfer.
\end{itemize}

%% file: sections/related_works.tex
\section{Related Works}

\subsection{On-Policy Distillation}

On-policy distillation (OPD) combines on-policy sampling with dense supervision: the student generates trajectories, and the teacher scores each token on student-visited states \cite{agarwal2024policy,lu2025opd}. OPD typically minimizes the reverse KL, which is mode-seeking and avoids forcing a lower-capacity student to model the teacher's long tail \cite{gu2024minillm}. Its gradient is equivalent to a policy-gradient update with dense token-level reward
$r_t=\log\pi_T(a_t\mid s_t)-\log\pi_\theta(a_t\mid s_t)$, or, equivalently, KL-constrained RL with teacher reward $\log\pi_T(a_t\mid s_t)$ and unit KL weight \cite{yang2026learning,zhao2026decoupling}. This dense, on-policy supervision improves sample efficiency, reduces distribution shift by training on states visited by the student, and mitigates catastrophic forgetting associated with off-policy SFT \cite{yang2025qwen3,agarwal2024policy,shenfeld2026self}. Related self-distillation methods similarly acquire new skills while preserving prior capabilities without explicit rewards \cite{zhao2026self,hubotter2026reinforcement}.

OPD has therefore become a common component of production post-training pipelines. Qwen3 \cite{yang2025qwen3} first uses off-policy distillation to establish reasoning capabilities and then OPD for logit alignment. MiMo-V2-Flash further scales this approach through multiple domain-specialized RL teachers, enabling a 309B/15B-active student to match models with 2--3$\times$ more total parameters \cite{xiao2026mimo}. However, OPD relies on a sufficiently stronger teacher: because its reward is derived from teacher log-probabilities on student trajectories, the signal becomes uninformative when teacher and student policies are too similar \cite{li2026rethinking}.

\subsection{Relative Policy Differences in Post-Training}

As mentioned above, OPD remains informative only when the teacher is sufficiently stronger than the student, since it directly uses the teacher's log-probability as reward \cite{li2026rethinking,wang2026demystifying}. Recent work relaxes this requirement by transferring how a teacher improves rather than its final policy. Given a reference policy $\pi_{\mathrm{ref}}$ and its adapted expert $\pi_{\exp}$, these methods use the log-ratio
$\log \pi_{\exp}(a\mid s)-\log \pi_{\mathrm{ref}}(a\mid s)$
on student rollouts as a dense learning signal. The student thus follows a direction of improvement instead of matching an absolute distribution. Existing methods construct the reference--expert pair through instruction tuning \cite{heo2026policy}, reinforcement learning on a weaker model \cite{feng2026weak}, or reward-based optimization of an auxiliary proxy policy \cite{fu2026proxyopd}. Because only the policy difference is transferred, the source model need not outperform the student, allowing weaker, older, or independently trained models to provide useful supervision.

Despite promising results across domains, prior work provides limited understanding of when and across which models relative policy transfer succeeds. Its effectiveness depends on both how the reference--expert pair is constructed and which target model receives the resulting policy shift. Existing studies largely evaluate specific pairs on students from the same model generation and at similar scales. It remains unclear which relative policies are useful, whether they transfer across models with different priors or capability levels, and when their benefits vanish. We systematically vary both the source policy pair and the target model to characterize effective relative policies and the boundaries of their transferability.

%% file: sections/method.tex
\section{A Unified Policy Optimization View of OPD}
\subsection{Conventional OPD as KL-Regularized Optimization}
\label{sec:opd_as_kl_regularized_optimization}

Conventional OPD trains a target policy
$\pi_\theta$ to match a fixed teacher policy $\pi_{\mathrm{T}}$ on states
visited by the target itself~\cite{agarwal2024policy}. It minimizes the reverse
KL objective
\begin{equation}
\mathcal{L}_{\mathrm{OPD}}(\theta)
=
\mathbb{E}_{s\sim d_{\pi_\theta}}
\left[
D_{\mathrm{KL}}
\left(
\pi_\theta(\cdot\mid s)
\,\|\, 
\pi_{\mathrm{T}}(\cdot\mid s)
\right)
\right]
\label{eq:opd_objective}
\end{equation}
where $d_{\pi_\theta}$ is the state distribution induced by target rollouts.
This objective performs \emph{absolute distribution matching}: at each
visited state, $\pi_\theta$ is directly optimized toward the distribution of
$\pi_{\mathrm{T}}$.

To expose the policy-optimization structure underlying this objective, let
$\pi_{\mathrm{ref}}$ denote the initial target policy. For each state $s$,
the negative reverse KL can be rewritten as
\vspace{-0.5em}
\begin{align}
&-
D_{\mathrm{KL}}
\left(
\pi_\theta(\cdot\mid s)
\,\|\,
\pi_{\mathrm{T}}(\cdot\mid s)
\right)
\nonumber=
\mathbb{E}_{a\sim\pi_\theta(\cdot\mid s)}
\left[
\log
\frac{\pi_{\mathrm{T}}(a\mid s)}
     {\pi_\theta(a\mid s)}
\right]
\nonumber\\
&=
\mathbb{E}_{a\sim\pi_\theta(\cdot\mid s)}
\left[
\log
\frac{\pi_{\mathrm{T}}(a\mid s)}
     {\pi_{\mathrm{ref}}(a\mid s)}
\right]
-
D_{\mathrm{KL}}
\left(
\pi_\theta(\cdot\mid s)
\,\|\,
\pi_{\mathrm{ref}}(\cdot\mid s)
\right)
\label{eq:opd_decomposition}
\end{align}

Consequently, minimizing Eq.~\ref{eq:opd_objective} is equivalent to
maximizing
\begin{equation}
J_{\mathrm{OPD}}(\theta)
=
\mathbb{E}_{s\sim d_{\pi_\theta}}
\left[
\mathbb{E}_{a\sim\pi_\theta(\cdot\mid s)}
\left[
\log
\frac{\pi_{\mathrm{T}}(a\mid s)}
     {\pi_{\mathrm{ref}}(a\mid s)}
\right]
-
D_{\mathrm{KL}}
\left(
\pi_\theta(\cdot\mid s)
\,\|\,
\pi_{\mathrm{ref}}(\cdot\mid s)
\right)
\right]
\label{eq:opd_policy_optimization}
\end{equation}

Eq.~\ref{eq:opd_policy_optimization} gives a KL-regularized policy
optimization view of conventional OPD: the relative log-difference
$\log\pi_{\mathrm{T}}-\log\pi_{\mathrm{ref}}$ acts as a reward-like signal
that drives the update, while the KL term anchors the policy to
$\pi_{\mathrm{ref}}$.

\vspace{-0.5em}
\subsection{From Absolute to Relative Matching: An Implicit Reward Perspective}
\label{sec:relative_policy_differences}
\vspace{-0.5em}
A key feature of conventional OPD is that the same reference policy appears
in both the reward-like term and the KL regularizer. These two occurrences
of $\pi_{\mathrm{ref}}$ cancel, reducing the objective to direct matching of
the teacher distribution. This coupling makes conventional OPD an absolute
matching method, which can be restrictive when the teacher and target have
poorly aligned output distributions~\cite{li2026rethinking}.

We instead consider whether the relative log-difference can serve as a transferable learning signal on its own. Let $(\pi_\phi^{+},\pi_\phi)$ be a \emph{proxy pair}, where, for each state $s$, the proxy expert $\pi_\phi^{+}$ is obtained from the proxy reference $\pi_\phi$ by solving a one-step KL-regularized optimization problem with token-level task reward $r^{*}(s,a)$:
\vspace{-0.5em}
\begin{equation}
\pi_\phi^{+}
=
\arg\max_{\pi}
\left\{
\mathbb{E}_{a\sim\pi(\cdot\mid s)}
\left[r^{*}(s,a)\right]
-
\tau
D_{\mathrm{KL}}
\left(
\pi(\cdot\mid s)
\,\|\,
\pi_\phi(\cdot\mid s)
\right)
\right\},
\label{eq:proxy_reward_optimization}
\end{equation}
where $\tau>0$ controls the strength of KL regularization. We define the \emph{relative policy difference} of this pair as
\vspace{-1.25em}
\begin{equation}
\Delta_\phi(s,a)
\triangleq
\log
\frac{\pi_\phi^{+}(a\mid s)}
     {\pi_\phi(a\mid s)}.
\label{eq:relative_policy_difference}
\end{equation}

The optimal solution to Eq.~\ref{eq:proxy_reward_optimization} satisfies
\begin{equation}
\Delta_\phi(s,a)
=
\frac{1}{\tau}r^{*}(s,a)
-
\log Z_\phi(s),
\qquad
Z_\phi(s)
=
\sum_{a'}
\pi_\phi(a'\mid s)
\exp\left(
\frac{r^{*}(s,a')}{\tau}
\right).
\label{eq:proxy_implicit_reward}
\end{equation}
A complete derivation is provided in Appendix~\ref{app:implicit_reward_derivation}.
Since $\log Z_\phi(s)$ depends only on the state, it cancels when
comparing actions. Thus, $\Delta_\phi$ preserves the action-wise reward
differences induced by $r^{*}$, up to the positive scale $1/\tau$, and
provides an implicit reward parameterization through the proxy policy
difference~\cite{rafailov2023direct}.

Importantly, $\Delta_\phi$ is defined entirely by the proxy pair and does
not depend on the target reference. This separates the source of the learning signal from the policy to which it is applied: the proxy pair $(\pi_\phi^{+},\pi_\phi)$ specifies the relative update, while a separate
target reference specifies where that update is transferred. We formalize
this relative matching procedure in the next subsection.
\vspace{-0.5em}
\subsection{Any-to-Any Policy Transfer}
\label{sec:any_to_any_transfer}
\vspace{-0.25em}
We now apply the relative policy difference $\Delta_\phi$ to a target policy
$\pi_\theta$ initialized from a reference $\pi_{\mathrm{ref}}$, where
$\pi_{\mathrm{ref}}$ need not coincide with the proxy reference $\pi_\phi$.
The target is optimized by maximizing

\begin{equation}
\mathcal{J}_{\mathrm{A^2}}(\theta)
=
\mathbb{E}_{s\sim d_{\pi_\theta}}
\left[
\mathbb{E}_{a\sim\pi_\theta(\cdot\mid s)}
\left[
\alpha\,\Delta_\phi(s,a)
\right]
-
\beta
D_{\mathrm{KL}}
\left(
\pi_\theta(\cdot\mid s)
\,\|\,
\pi_{\mathrm{ref}}(\cdot\mid s)
\right)
\right],
\label{eq:any_to_any_objective}
\end{equation}

where $\alpha>0$ scales the transferred signal and $\beta>0$ regularizes the target toward its reference. The first term follows the action preferences encoded by the proxy update.


Eq.~\ref{eq:any_to_any_objective} realizes \emph{relative
distribution matching}. Instead of matching the absolute distribution of
$\pi_\phi^{+}$, the target keeps its own reference $\pi_{\mathrm{ref}}$ and
reweights it by the relative change $\pi_\phi^{+}/\pi_\phi$, with the ratio
$\alpha/\beta$ controlling the transfer strength.

Conventional OPD is recovered from Eq.~\ref{eq:any_to_any_objective} by
setting $\pi_\phi^{+}=\pi_{\mathrm{T}}$, $\pi_\phi=\pi_{\mathrm{ref}}$, and $\alpha=\beta$. When the proxy and target references differ, their cancellation disappears and only the relative policy difference is transferred. We call this generalization \emph{any-to-any policy transfer} ($\mathrm{A}^2\text{-OPD}$): the two ``any''s denote an independently chosen proxy pair and target, respectively. This permits arbitrary source--target combinations in principle, although not all transfer effectively. Our experiments characterize which proxy differences are useful and which targets can realize them.

Relative policy update transfer is an active research direction~\cite{feng2026weak, fu2026proxyopd, heo2026policy}. Our formulation unifies these methods under a broader policy-optimization view. As shown in Table~\ref{tab:special_cases}, they mainly differ in the proxy pair, target reference, and transfer strength.

\begin{table}[h]
\centering
\small
\caption{
Existing OPD methods as special cases of any-to-any policy transfer
(Eq.~\ref{eq:any_to_any_objective}).
Here, $T^+$, $T$, and $S$ denote the teacher expert, teacher base,
and student base, respectively.
}
\label{tab:special_cases}
\renewcommand{\arraystretch}{1.35}
\begin{tabular}{lccccc}
\toprule
Method
& $\pi_\phi^{+}$
& $\pi_\phi$
& $\pi_{\mathrm{ref}}$
& $\alpha/\beta$
& Advantage\\
\midrule
OPD
& $T^+$
& $S$
& $S$
& $1$
& $\log \pi_{T^+} - \log \pi_\theta$ \\

ExOPD
& $T^+$
& $S$
& $S$
& $\lambda$
& $\lambda(\log \pi_{T^+} - \log \pi_S)
   - (\log \pi_\theta - \log \pi_S)$ \\

Relative Policy Transfer
& $T^+$
& $T$
& $S$
& $\lambda$
& $(\log \pi_{T^+} - \log \pi_T)
   - \lambda(\log \pi_\theta - \log \pi_S)$ \\
\bottomrule
\end{tabular}
\end{table}

Follow the prior works~\cite{fu2026proxyopd,li2026rethinking},
we optimize the objective $\mathcal{J}_{\mathrm{A^2}}$ directly through its policy gradient, using samples drawn from the target policy itself.

Let $\ell_\theta(s,a) = \log\frac{\pi_\theta(a\mid s)}{\pi_{\mathrm{ref}}(a\mid s)}$ denote the log-ratio between the target policy and its reference. Here $\Delta_\phi$ is fixed during target training, while $\ell_\theta$ measures the current deviation of $\pi_\theta$ from $\pi_{\mathrm{ref}}$.

Differentiating Eq.~\ref{eq:any_to_any_objective} gives
\begin{equation}
\nabla_\theta \mathcal{J}_{\mathrm{A^2}}(\theta)
=
\mathbb{E}_{s\sim{d}_{\pi_\theta},\,a\sim\pi_\theta(\cdot\mid s)}
\left[
u_\theta(s,a)\,
\nabla_\theta \log\pi_\theta(a\mid s)
\right],
\qquad
u_\theta(s,a)
=
\alpha\,\Delta_\phi(s,a)
-
\beta\,\ell_\theta(s,a),
\label{eq:A2_policy_gradient}
\end{equation}

where $d_{\pi_\theta}$ is the state distribution induced by target
rollouts. Although $\ell_\theta$ depends on $\theta$, its explicit derivative
contributes
$-\beta\,\mathbb{E}_{a\sim\pi_\theta}[\nabla_\theta\log\pi_\theta(a\mid s)]=0$,
so Eq.~\ref{eq:A2_policy_gradient} is exactly the gradient of the
KL-regularized objective in the standard score-function form with per-action
weight $u_\theta$.

The weight $u_\theta$ still carries a state-dependent offset inherited from
the proxy pair. Substituting the implicit-reward relation
$\alpha\Delta_\phi(s,a)=\tfrac{\alpha}{\tau}r^{*}(s,a)-\alpha\log Z_\phi(s)$
from Eq.~\ref{eq:proxy_implicit_reward} gives
\begin{equation}
u_\theta(s,a)
=
\underbrace{
\frac{\alpha}{\tau}\,r^{*}(s,a)
-
\beta\,\ell_\theta(s,a)
}_{\text{action-dependent}}
\;-\;
\underbrace{
\alpha\log Z_\phi(s)
}_{\text{action-independent}}.
\label{eq:A2_score_decomposition}
\end{equation}
The offset $\alpha\log Z_\phi(s)$ is constant across actions and involves the
intractable normalizer $Z_\phi(s)$. To remove it, we center $u_\theta$ over
actions,
\vspace{-0.5em}
\begin{equation}
A_\theta(s,a)
=
u_\theta(s,a)
-
\mathbb{E}_{a'\sim\pi_\theta(\cdot\mid s)}
\left[
u_\theta(s,a')
\right]
\label{eq:A2_centered_score}
\end{equation}
so that $\alpha\log Z_\phi(s)$ cancels between $u_\theta$ and the baseline.
The centered score thus depends only on the action-dependent part of
Eq.~\ref{eq:A2_score_decomposition} and can be computed directly from
$\Delta_\phi$ and $\ell_\theta$, without forming $Z_\phi(s)$. Because the
subtracted baseline does not depend on the sampled action, replacing
$u_\theta$ with $A_\theta$ leaves the expected gradient in
Eq.~\ref{eq:A2_policy_gradient} unchanged while reducing its variance.

The resulting on-policy actor loss is
\begin{equation}
\mathcal{L}(\theta)
=
-
\mathbb{E}_{s\sim\rho_{\pi_\theta},\,a\sim\pi_\theta(\cdot\mid s)}
\left[
\operatorname{sg}\!\left(
A_\theta(s,a)
\right)
\log\pi_\theta(a\mid s)
\right]
\label{eq:A2_actor_loss}
\end{equation}

where $\operatorname{sg}(\cdot)$ stops gradients through the centered score,
so that differentiating $\mathcal{L}(\theta)$ recovers
Eq.~\ref{eq:A2_policy_gradient}. For autoregressive language models,
$s_t=(x,y_{<t})$ and $a_t=y_t$, so $\Delta_\phi$ and $\ell_\theta$ are
evaluated token-wise along trajectories sampled from the current target
policy.

%% file: sections/experiment.tex
\vspace{-0.5em}
\section{When Do Relative Policy Differences Transfer?}
\label{sec:transfer_analysis}
\vspace{-0.5em}
\subsection{Experiment Setup and Metrics}
\label{sec:experiment_setup}
\vspace{-0.5em}
\paragraph{Protocol.}
All experiments optimize the same objective (Eq.~\ref{eq:A2_actor_loss}) with a fixed transfer strength $\alpha/\beta=1$. Models are drawn from the Qwen3 and Qwen2.5 families. 
For efficiency, all experiments use the top-$16$ token distribution instead of the full vocabulary~\cite{li2026rethinking}.
\vspace{-0.5em}
\paragraph{Training and evaluation.}
We follow the prior work~\cite{fu2026proxyopd}, proxy pairs are obtained by RL or SFT on DeepMath-103K~\cite{he2025deepmath} for math and Eurus-RL-Code~\cite{cui2025process} for code, using a ruled verifiable reward. Sampling uses temperature $1.0$ and top-$p$ $1.0$ for both training and evaluation. We evaluate on eight math benchmarks (AIME24/25/26, SMT25, CMIMC25, HMMT25-Feb/Nov, HMMT26-Feb) with 16 samples, and on three code benchmarks (HumanEval+, MBPP+, LCB) with 8 samples. Model details are in Appendix~\ref{app:experiment_setup}.
\vspace{-0.5em}
\paragraph{Metrics.}
Let $P(\pi)$ denote the average task performance of policy $\pi$. We measure the proxy pair capability gain and the target model improvement as
\begin{equation}
G_P=P(\pi_\phi^{+})-P(\pi_\phi),\qquad
G_T=P(\pi^*_\theta)-P(\pi_{\mathrm{ref}}),
\label{eq:transfer_gains}
\end{equation}
and report $R_{\mathrm{perf}}=G_T/G_P$ to compare their scales. These metrics describe the policy difference and its transfer in performance space. We further use KLd diagnostics to characterize transfer beyond performance differences. $K_\text{Proxy}$ quantifies the proxy expert--reference difference, and $K_\text{Base}$ the proxy--target reference gap. $K_\text{Target}$ quantifies the difference between the target reference and the optimized target. $K_\text{Induced}$ quantifies the difference between $K_\text{Target}$ and the distribution induced by the proxy log-policy difference. Formal definitions are provided in
Appendix~\ref{app:metrics}.


\begin{figure*}[!h]
    \centering
    \includegraphics[width=\textwidth]{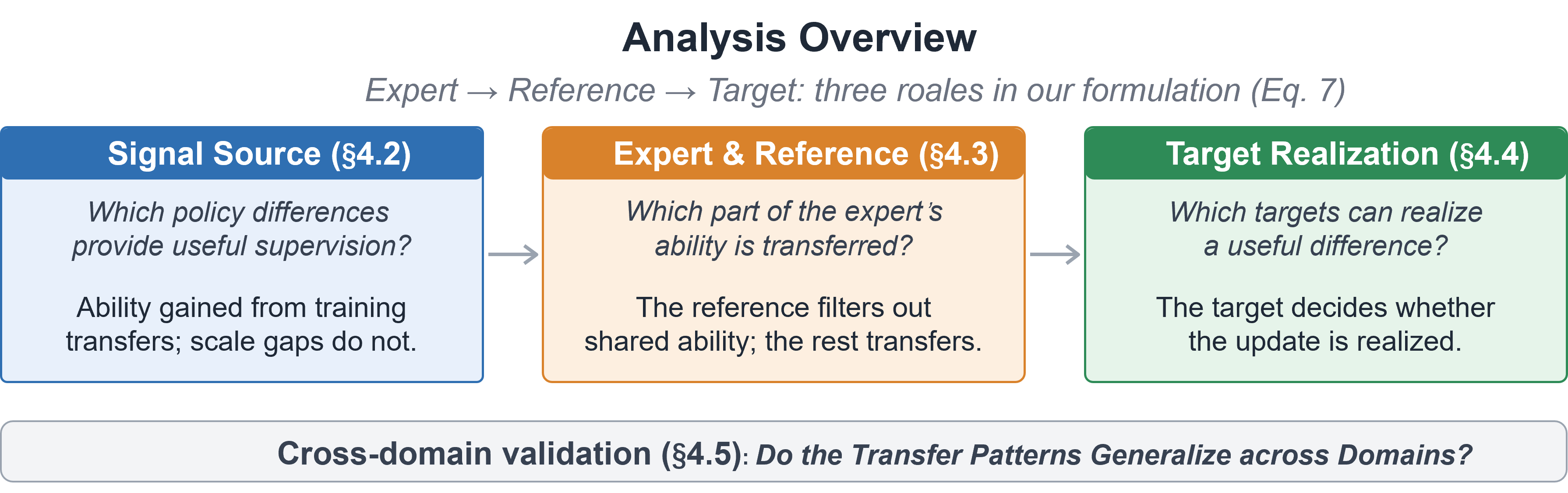}
    \label{fig:analysis_roadmap}
\end{figure*}

\subsection{Which Policy Differences Provide Useful Supervision?}
\label{exp:transfer_sources}

A stronger proxy expert may differ from its reference for two main reasons: task-specific optimization, or simply larger scale. We ask whether these two sources of difference provide equally useful supervision. We fix Qwen3-8B as the target and compare nine proxy pairs constructed from RL and SFT updates, scale gaps, and one alternative reference.

\paragraph{Task-directed optimization creates transferable differences.}
Eq.~\ref{eq:proxy_reward_optimization} shows that KL-regularized task
optimization makes $\log\pi_\phi^{+}-\log\pi_\phi$ encode a task-relevant
update. More broadly, this interpretation applies approximately to experts obtained through rewardd RL or SFT with offline data, even without explicit KL regularization. In contrast, scale differences between independently trained models need not encode a task-relevant direction.

Table~\ref{tab:rq1_main} follows this split. All task-optimized differences improve the target: the RL and SFT pairs yield $G_T$ ranging from $9.78$ to $32.23$, and Qwen3-4B-RL with an alternative Qwen3-1.7B reference achieves $25.85$, showing that the reference need not be expert's training origin. In contrast, the four scale-only pairs yield much smaller target gains, despite positive proxy pair capability gaps.

\input{tables/rq1}

\begin{figure}[t]
    \centering
    \begin{subfigure}[t]{0.56\textwidth}
        \centering
        \includegraphics[width=\linewidth]
        {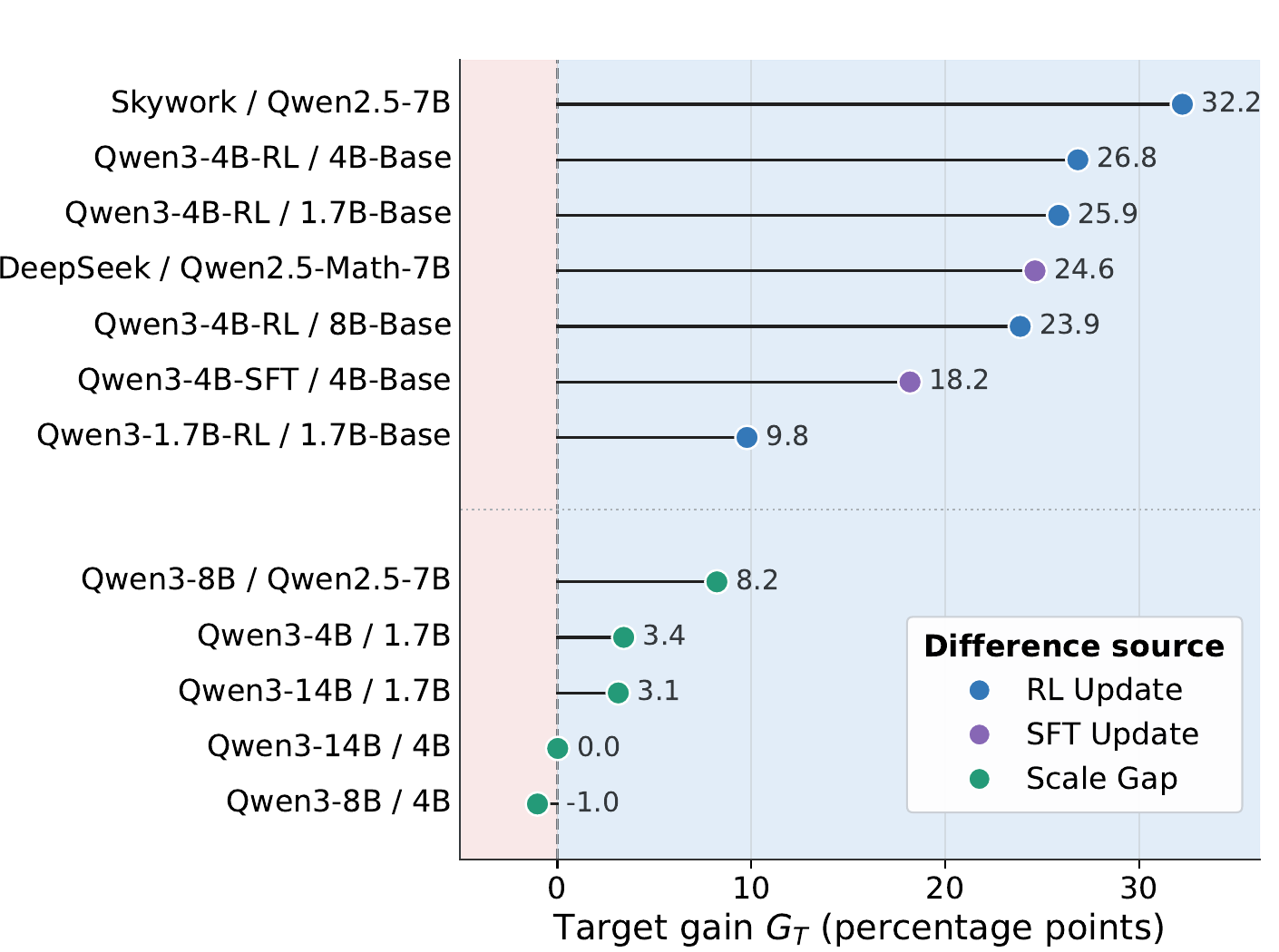}
        \caption{Target gains from different proxy pairs.}
        \label{fig:rq1_target_gain}
    \end{subfigure}
    \hfill
    \begin{subfigure}[t]{0.42\textwidth}
        \centering
        \includegraphics[
            width=\linewidth,
            trim={5pt 0pt 0pt 0pt},
            clip
        ]{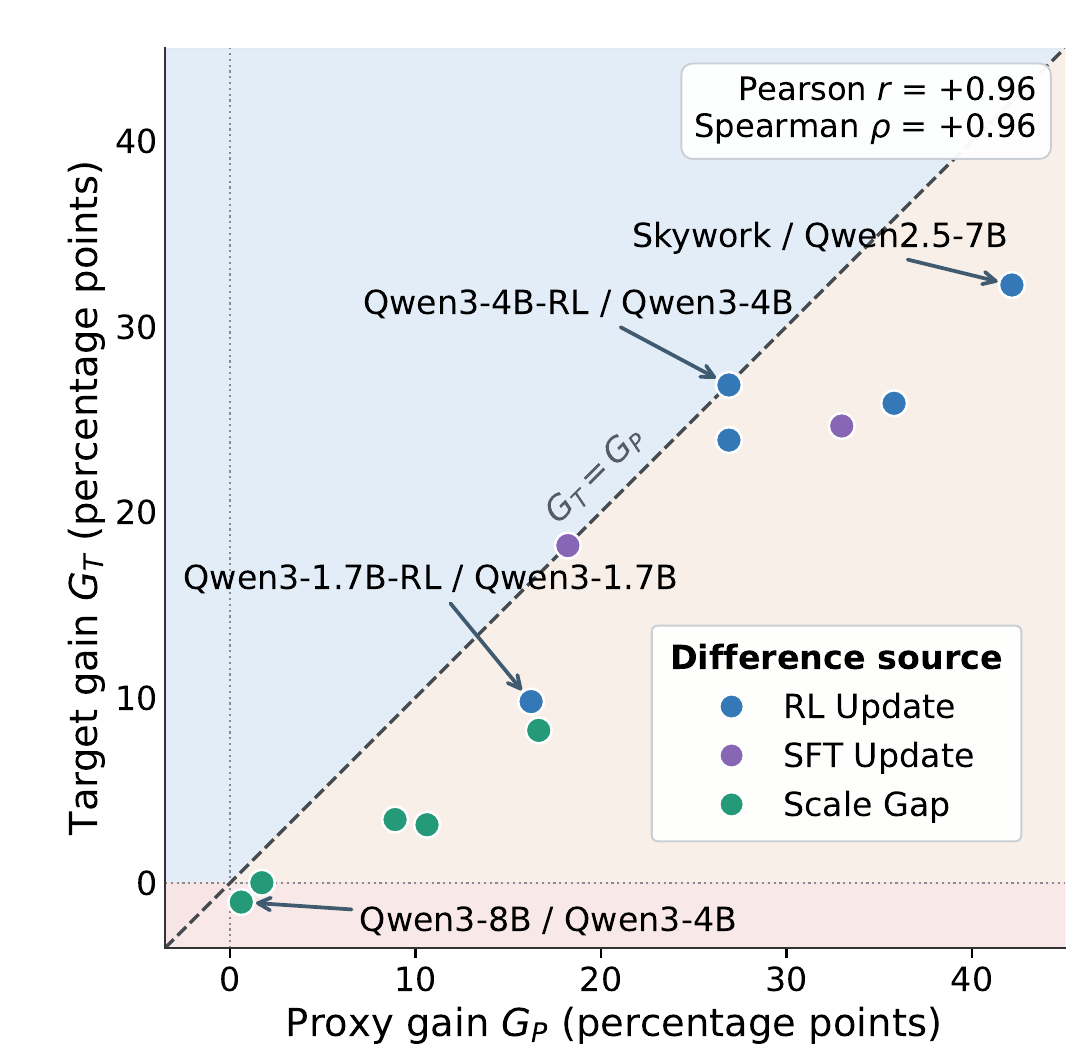}
        \caption{Proxy capability gain vs. target gain}
        \label{fig:rq1_gain_relation}
    \end{subfigure}
    \captionsetup{skip=2pt}
    \caption{
        \textbf{Transferability across sources of policy difference.}
    }
    \label{fig:rq1_transfer_sources}
    \vspace{-1.25em}
\end{figure}

\vspace{-0.25em}
\paragraph{Transfer depends on task relevance, not merely update magnitude.}
Across all pairs, the proxy gain $G_P$ is strongly associated with the target gain $G_T$ (Pearson $r=0.96$, Spearman $\rho=0.93$; Figure~\ref{fig:rq1_transfer_sources}). However, update magnitude $K_\text{Induced}$ alone is insufficient to explain transfer. With Qwen3-8B as the target and Qwen3-1.7B as the proxy reference, the task-optimized pair (Qwen3-4B-RL / Qwen3-1.7B) and the scale-only pair (Qwen3-4B / Qwen3-1.7B) induce updates of comparable magnitude ($K_I=0.3531$ vs.\ $0.4181$), but yield substantially different target gains ($25.85$ vs.\ $3.42$). This comparison indicates that transfer depends on the task-relevant direction encoded by policy difference, rather than its magnitude alone.

\noindent
\begin{minipage}[t]{0.44\linewidth}
\vspace{0pt}
\textbf{The target follows a task-trained difference but not a scale-only one.} Tracking the token-level overlap between target and proxy expert for these two pairs (Figure~\ref{fig:rq1_overlap}) shows the mechanism directly. Although the scale-only pair starts higher, the target's overlap with the task-trained expert rises from $0.75$ to about $0.81$ and overtakes it within $20$ steps, while the scale-only overlap drifts down to $0.78$. These contrasting trajectories suggest that the target follows the task-aligned direction characterized by Eq.~\ref{eq:proxy_reward_optimization}, consistent with its larger $G_T$.
\end{minipage}
\hfill
\begin{minipage}[t]{0.53\linewidth}
\vspace{-1.5em}
\centering
\includegraphics[width=\linewidth]{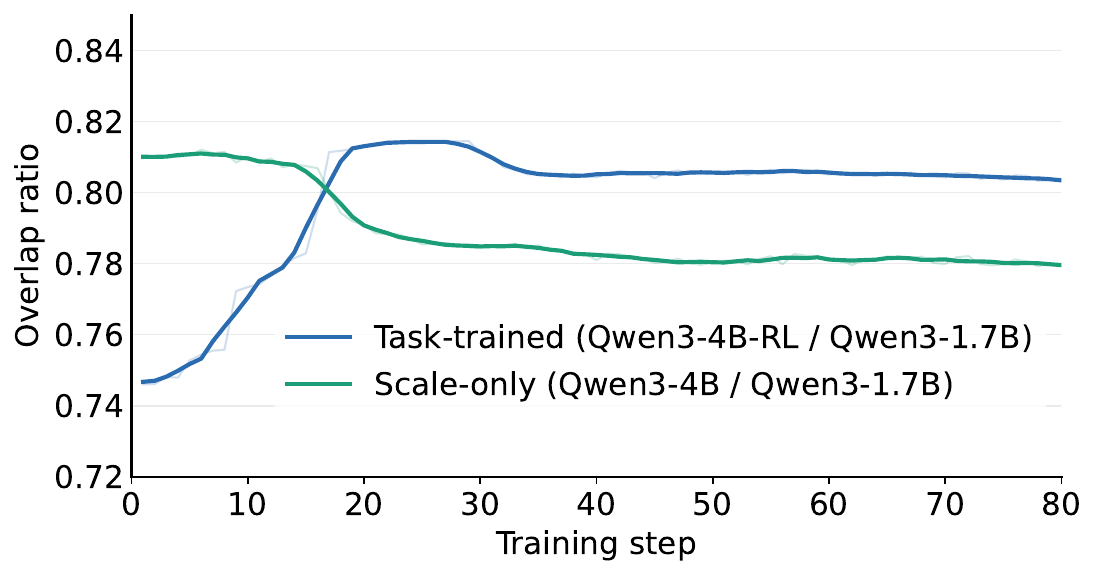}
\captionsetup{skip=2pt}
\captionof{figure}{Overlap ratio between the target and proxy expert
during training.}
\label{fig:rq1_overlap}
\end{minipage}
\vspace{0.8em}
\begin{takeaway}
\small
\textbf{A stronger model is not automatically a better teacher.}
Improvements from optimizing toward an explicit or implicit task reward, whether by RL or SFT, transfer consistently, while improvements from increasing model size alone transfer weakly or unreliably.
\end{takeaway}

\subsection{Which Part of the Expert's Capability Gets Transferred?}
\label{exp:expert_reference}
\vspace{-1em}
A policy difference is jointly defined by proxy expert and proxy reference. Section~\ref{exp:transfer_sources} showed that the difference must encode a task-optimized capability; here we ask \textit{how the expert and reference each shape that capability}. We fix Qwen3-8B as the target, and study two controlled sweeps: fixing the reference to Qwen3-4B while varying the expert, and fixing the expert to Qwen3-4B-RL while varying the reference.

\input{tables/rq2}
\vspace{-1em}
\paragraph{The expert sets how much capability is available.}
With the reference fixed to Qwen3-4B, moving the expert from the early to
the regular and late RL checkpoints raises $G_P$ from $22.09$ to $26.89$ and
$32.28$, and $G_T$ tracks this trend ($22.42\rightarrow26.84\rightarrow
28.18$; Table~\ref{tab:rq2_main}). A stronger expert thus makes more
capability available, although the additional capability is only partly realized
as the expert grows stronger ($R_{\mathrm{perf}}$ falls from $1.015$ to $0.873$).
\vspace{-1em}
\paragraph{The reference isolates the residual capability for transfer.}
Fixing the expert to Qwen3-4B-RL and changing the reference shows the complementary role. Replacing Qwen3-4B with the early RL checkpoint cuts $G_T$ from $26.84$ to $9.62$, because the difference now exposes only the increment from the early to the regular checkpoint, rather than the full increment over the base. The reference therefore acts as a filter: it removes the capability already shared with the expert and leaves only the residual for transfer. This filter need not be the expert's training origin. Using Qwen2.5-3B as the reference gives the largest target gain ($29.09$), despite the largest $K_\text{Base}=1.23$.
\vspace{-1em}

\begin{figure*}[t]
    \centering
    \includegraphics[width=\textwidth]{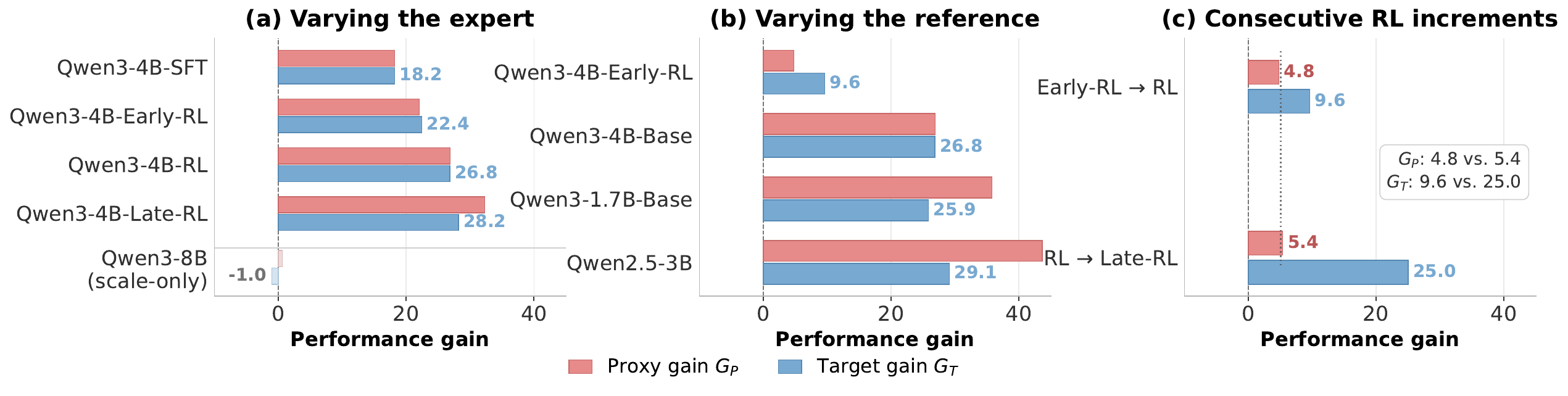}
    \vspace{-1em}
    \captionsetup{skip=2pt}
    \caption{\textbf{Effects of the proxy expert and reference on transfer.}
    Qwen3-8B is fixed as the target.
    }
    \vspace{-1.5em}
\end{figure*}

\paragraph{Token-level overlap reveals the residual selected by the reference.}
To directly examine the filtering effect, we measure the top-16 overlap ratio $\mathcal{O}_{E,R}$ between the proxy expert and proxy reference on the same fixed set of target states. With Qwen3-4B-RL fixed as the expert, $\mathcal{O}_{E,R}$ decreases from $0.916$ with the early-RL reference to $0.839$ with Qwen3-4B and $0.625$ with Qwen2.5-3B. The corresponding target gains increase from $9.62$ to $26.84$ and $29.09$. Thus, a reference that shares more high-probability tokens with expert removes more of the expert's policy change, leaving a smaller residual for transfer. This pattern is not merely a restatement of the association between $G_P$ and
$G_T$ observed in Section~\ref{exp:transfer_sources}. The early-to-regular and regular-to-late RL increments have similar proxy gains ($G_P=4.80$ and $5.39$),
but different expert--reference overlaps ($0.916$ and $0.869$), and produce a much larger target gain ($25.03$ vs. $9.62$). Therefore, similar target performance gains can encode substantially different policy-space residuals. The reference determines not only how much proxy performance differs, but also the structure and strength of the update presented to the target.


\begin{takeaway}
\small
The expert supplies the capability, while the reference filters out what the two models already share, leaving the remaining difference for transfer.
\end{takeaway}

\subsection{Which Targets Can Realize a Useful Difference?}
\label{sec:rq3}
\vspace{-0.5em}
Sections~\ref{exp:transfer_sources} and~\ref{exp:expert_reference} identified
which policy differences carry useful capability. A difference helps only if the target can realize it. We fix proxy pairs that already transfer well and vary the target: four Qwen3d proxy pairs, each applied to six targets (Qwen3-1.7B/4B/8B and Qwen2.5-7B/7B-Instruct/7B-Math). We report the results in the Table~\ref{tab:rq3_main}).

\begin{figure*}[t]
    \centering
    \includegraphics[width=\textwidth]{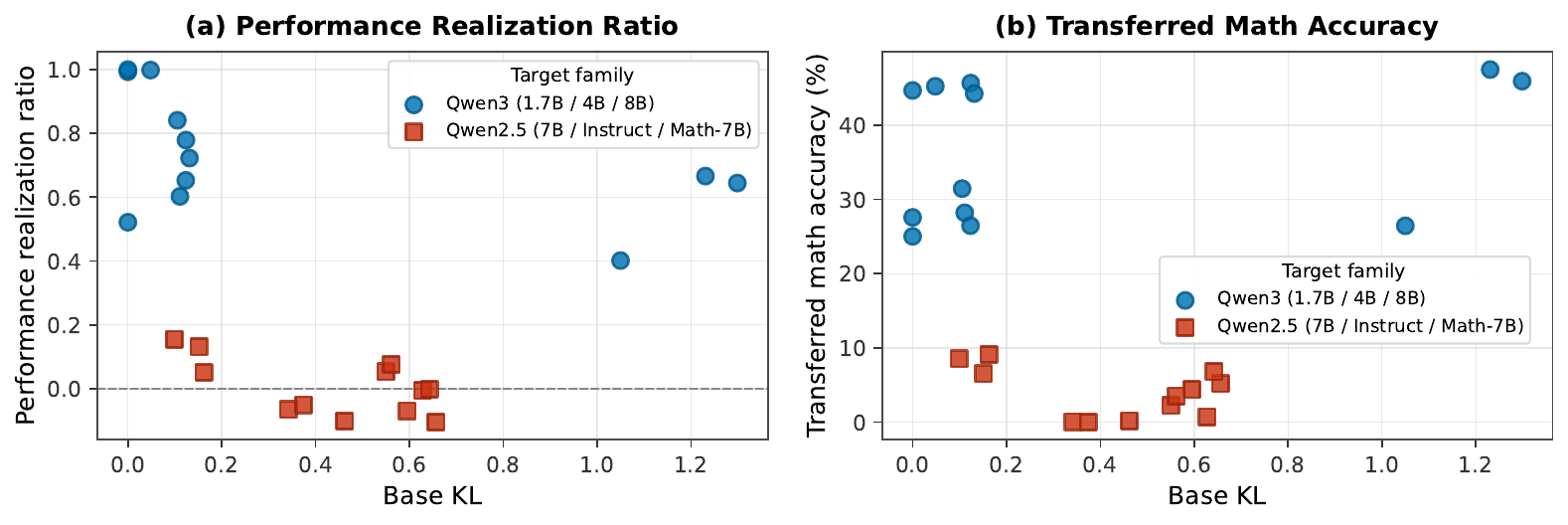}
    \vspace{-1.5em}
    \caption{\textbf{The outcome is not a function of policy distance.}
    Performance realization ratio (left) and transferred math accuracy (right)
    against the proxy--target base KL.}
    \label{fig:rq3_realization_vs_kl}
\end{figure*}

\input{tables/rq3}
\vspace{-1em}
\paragraph{The same difference helps some targets but not others.}
Applying one difference to different targets gives sharply different outcomes.
Qwen3 targets gain ranging from $9.78$ to $29.09$,
whereas Qwen2.5 targets under a Qwen3 reference gain almost nothing ($G_T\in[-2.46,\,2.72]$). Whether a difference helps is thus a property of the target, not of the difference alone. A natural assumption is that a target benefits when it is close to the proxy pair, but Figure~\ref{fig:rq3_realization_vs_kl} rejects this: neither $R_{\mathrm{perf}}$ nor transferred accuracy is monotone in the $K_\text{Base}$, and Qwen3 targets gain even at the largest $K_\text{Base}$ ($R_{\mathrm{perf}}\approx0.65$ at $K_\text{Base}>1.2$). The outcome splits by model family, not by distance.

\paragraph{Generation compatibility, not the difference, controls the outcome.}
To test whether the family split is causal, we hold the difference fixed and change only the reference's family. Replacing the Qwen3 reference with a Qwen2.5
one (the Qwen3-4B-RL / Qwen2.5-3B pair) turns the Qwen2.5 targets from near-zero or negative gains into positive ones ($G_T=6.74/2.25/5.76$), while Qwen3 targets stay successful (Table~\ref{tab:rq3_main}). This intervention shows that transfer failure is not an intrinsic limitation of the target or the useful difference. Instead, the reference determines whether the difference is presented in a form that the target can learn and generalize from; matching the model family is one effective choice in our experiments.

\begin{figure}[t]
    \centering
    \begin{subfigure}{0.49\linewidth}
        \centering
        \includegraphics[width=\linewidth]{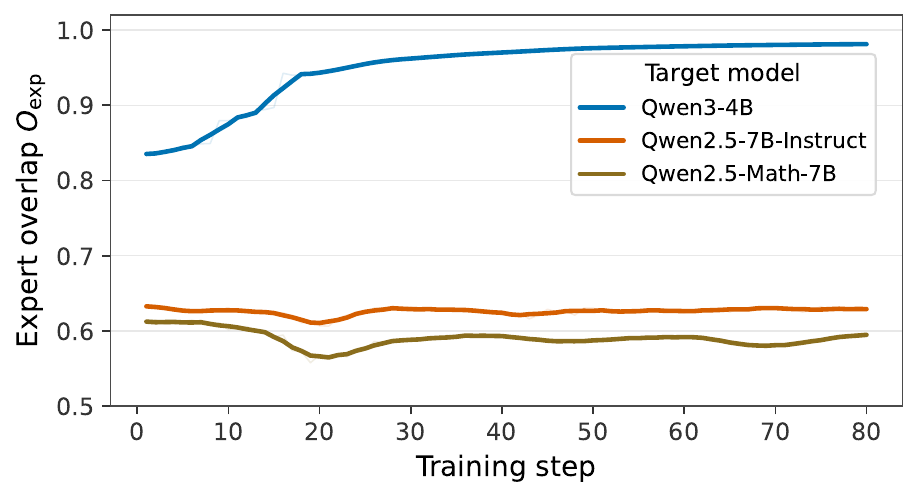}
        \caption{Overlap with the proxy expert}
        \label{fig:rq3_expert_overlap_by_target}
    \end{subfigure}
    \hfill
    \begin{subfigure}{0.49\linewidth}
        \centering
        \includegraphics[width=\linewidth]{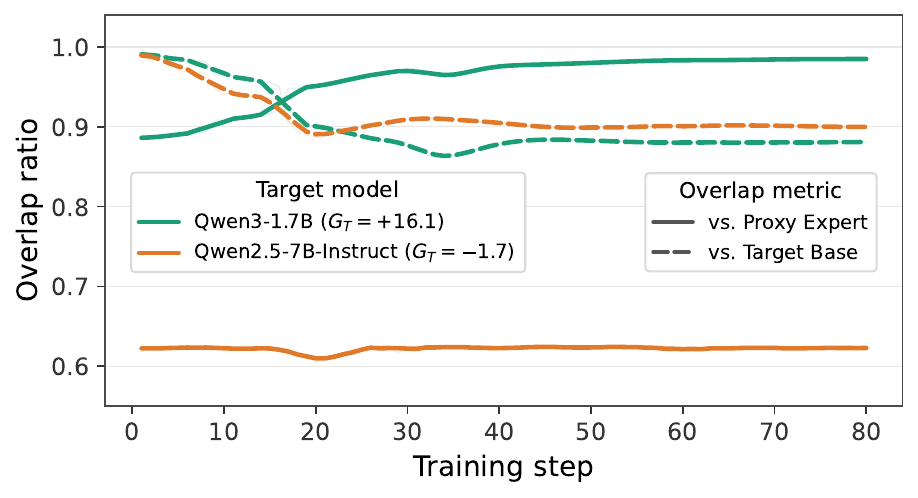}
        \caption{Overlap with expert vs.\ target base}
        \label{fig:rq3_overlap_expert_vs_base}
    \end{subfigure}
    \captionsetup{skip=2pt}
    \caption{\textbf{Benefiting targets move toward the proxy expert}(fixed Qwen3-1.7B proxy pair).}
    \label{fig:rq3_overlap}
\end{figure}

\begin{figure}[t]
    \centering
    \begin{subfigure}{0.345\linewidth}
        \centering
        \includegraphics[width=\linewidth]{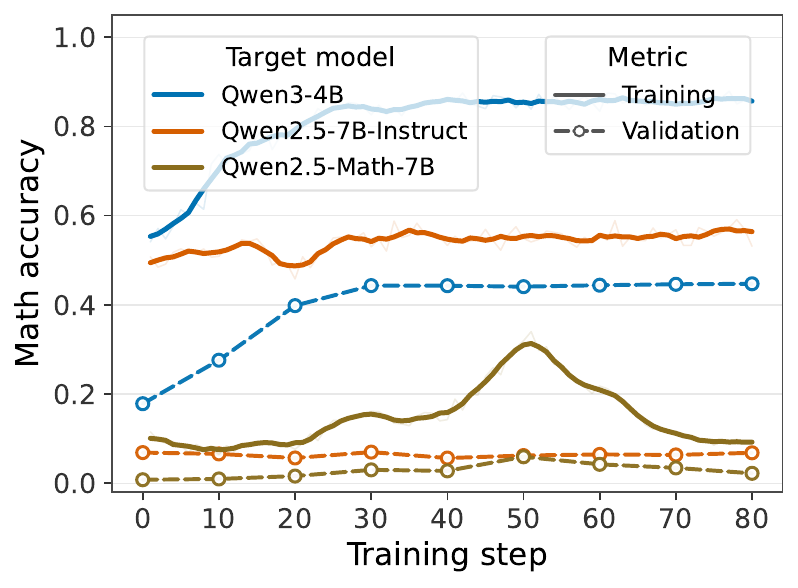}
        \caption{Train vs.\ validation across targets}
        \label{fig:rq3_trainval_by_target}
    \end{subfigure}
    \hfill
    \begin{subfigure}{0.635\linewidth}
        \centering
        \includegraphics[width=\linewidth]{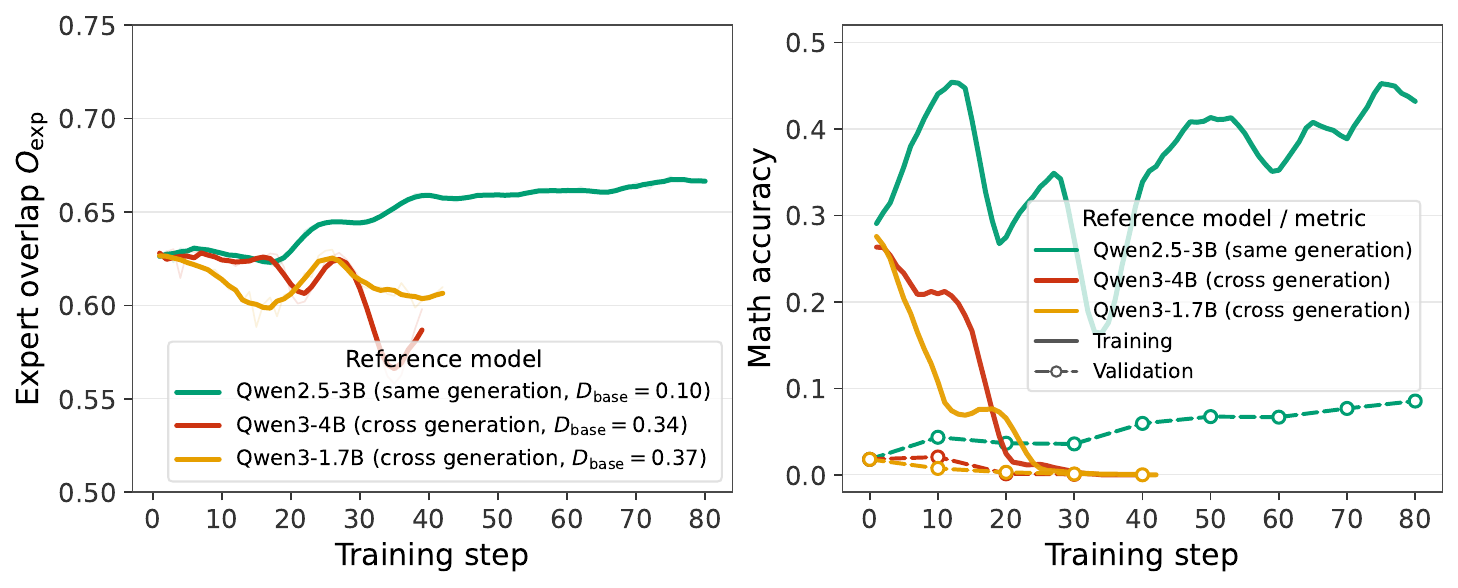}
        \caption{Effect of the reference family}
        \label{fig:rq3_reference_family_effect}
    \end{subfigure}
    \captionsetup{skip=2pt}
    \caption{\textbf{Failure is not undertraining, and matching the reference generation recovers the benefit.}}
    \label{fig:rq3_train_val}
\end{figure}

\paragraph{Benefiting means moving toward the expert and generalizing.}
Two diagnostics explain what compatibility buys. First, benefiting targets move toward the expert: for a fixed Qwen3-1.7B proxy pair, the Qwen3 target's overlap with the proxy expert rises steadily while the Qwen2.5 targets stay flat (Figure~\ref{fig:rq3_overlap}a); tracking both references confirms the direction, as the benefiting target ($G_T=+16.1$) moves toward the expert and away from its own base, whereas the failing target ($G_T=-1.7$) moves toward neither (Figure~\ref{fig:rq3_overlap}b). Second, this movement must generalize: for the incompatible Qwen2.5-Math-7B target, training accuracy rises and even spikes but validation accuracy stays near zero, so the update fits the training problems without transferring (Figure~\ref{fig:rq3_train_val}a). Matching the reference generation restores both: overlap with the expert grows and validation accuracy rises, while cross-generation references make both collapse (Figure~\ref{fig:rq3_train_val}b). Failure is therefore not undertraining but an inability to move toward the expert in a way that generalizes.

\begin{takeaway}
\small
\textbf{A useful policy difference is not automatically transferable to every
target.}
Transfer succeeds when the induced update moves the target toward the proxy
expert and generalizes beyond the training examples. Raw policy distance does
not predict this behavior, while choosing an appropriate reference can recover
an otherwise failed transfer.
\end{takeaway}

\subsection{Do the Transfer Patterns Generalize across Domains?}
\label{sec:cross_domain}

The preceding findings are established on mathematical reasoning tasks. We next examine whether the same qualitative transfer patterns appear in code generation. We select representative successful and failed configurations and evaluate matched configurations on both math and code. For each setting, we report the target gain $G_T$ over the corresponding target
base in Table~\ref{tab:cross_domain}.

\input{tables/rq4}

\paragraph{The success and failure patterns persist across domains.}
Transfer between Qwen3 models is positive on both math
($+26.8$ and $+25.9$) and code ($+2.3$ and $+1.9$). By contrast, transferring a Qwen3-derived difference to a Qwen2.5 target fails in both domains, producing gains of $-1.7$ and $-6.4$, respectively. Settings with an uninformative also yield no useful target improvement in either domain. Finally, the Qwen2.5-based proxy pair transfers successfully to Qwen3-8B on both math ($+24.6$) and code ($+9.1$), showing that the cross-domain consistency is not restricted to same-generation transfer. Although the gains are generally smaller on code, the signs and relative ordering of the tested configurations remain consistent across the two domains.

\begin{takeaway}
\small
\textbf{Transfer patterns persist across tasks:} configurations that succeed or fail on math show the same qualitative outcomes on code.
\end{takeaway}

%% file: tables/rq1.tex
\begin{table}[h]
    \centering
    \scriptsize
    \setlength{\tabcolsep}{6.8pt}
    \caption{
    Transfer results for policy differences from different sources, with Qwen3-8B fixed as the target. $^{\dagger}$Alternative reference: the proxy reference is not the expert's training origin.
    }
    \label{tab:rq1_main}
    \begin{tabular}{llccc}
        \toprule
       \makecell[c]{Source}
        & \makecell[c]{Proxy expert / reference}
        & \makecell[c]{Proxy performance\\($G_P$)}
        & \makecell[c]{Target performance\\($G_T$)}
        & \makecell[c]{$K_{\mathrm{Induced}}$} \\
        \midrule

        \multicolumn{5}{l}{\emph{Task-optimized differences}} \\

        RL
        & Qwen3-1.7B-RL / Qwen3-1.7B
        & $8.92\!\rightarrow\!25.15$ ($+16.23$)
        & $18.43\!\rightarrow\!28.21$ ($+9.78$)
        & 0.0773 \\

        RL
        & Qwen3-4B-RL / Qwen3-4B
        & $17.82\!\rightarrow\!44.71$ ($+26.89$)
        & $18.43\!\rightarrow\!45.27$ ($+26.84$)
        & 0.1066 \\

        RL
        & Skywork-7B / Qwen2.5-7B
        & $1.79\!\rightarrow\!43.93$ ($+42.14$)
        & $18.43\!\rightarrow\!50.66$ ($+32.23$)
        & 0.2292 \\

        RL$^{\dagger}$
        & Qwen3-4B-RL / Qwen3-1.7B
        & $8.92\!\rightarrow\!44.71$ ($+35.79$)
        & $18.43\!\rightarrow\!44.28$ ($+25.85$)
        & 0.3531 \\

        SFT
        & Qwen3-4B-SFT / Qwen3-4B
        & $17.82\!\rightarrow\!36.03$ ($+18.21$)
        & $18.43\!\rightarrow\!36.62$ ($+18.19$)
        & 0.1725 \\

        SFT
        & DeepSeek-R1-Distill-7B / Qwen2.5-Math-7B
        & $0.77\!\rightarrow\!33.74$ ($+32.97$)
        & $18.43\!\rightarrow\!43.06$ ($+24.63$)
        & 0.1785 \\

        \midrule
        \multicolumn{5}{l}{\emph{Scale-induced gaps}} \\

        Scale
        & Qwen3-4B / Qwen3-1.7B
        & $8.92\!\rightarrow\!17.82$ ($+8.90$)
        & $18.43\!\rightarrow\!21.85$ ($+3.42$)
        & 0.4181 \\

        Scale
        & Qwen3-8B / Qwen3-4B
        & $17.82\!\rightarrow\!18.43$ ($+0.61$)
        & $18.43\!\rightarrow\!17.41$ ($-1.03$)
        & 0.1379 \\

        Scale
        & Qwen3-14B / Qwen3-4B
        & $18.43\!\rightarrow\!19.54$ ($+1.11$)
        & $18.43\!\rightarrow\!18.45$ ($+0.02$)
        & 0.1730 \\

        Scale
        & Qwen3-8B / Qwen2.5-7B
        & $1.79\!\rightarrow\!18.43$ ($+16.64$)
        & $18.43\!\rightarrow\!26.66$ ($+8.23$)
        & 0.1522 \\

        \bottomrule
    \end{tabular}
\end{table}

%% file: tables/rq2.tex
\begin{table}[t]
    \centering
    \scriptsize
    \setlength{\tabcolsep}{4.2pt}
    \caption{
    \textbf{Effects of the proxy expert and reference on transfer}.
    The \colorbox{gray!10}{shaded} row is equivalent to standard OPD.
    Full policy-distance statistics are in Appendix~\ref{app:metrics}.
    }
    \label{tab:rq2_main}
    \begin{tabular}{llccccc}
        \toprule
        \makecell[c]{Proxy expert} 
        & \makecell[c]{Proxy reference}
        & \makecell[c]{Proxy performance\\($G_P$)}
        & \makecell[c]{Target performance\\($G_T$)}
        & \makecell[c]{$K_{\mathrm{Base}}$}
        & \makecell[c]{$R_{\mathrm{perf}}$}
        & \makecell[c]{$\mathcal{O}_{E,R}$} \\
        \midrule

        \multicolumn{7}{l}{
        \textit{Varying the expert; reference = Qwen3-4B-Base}}\\
        \addlinespace[2pt]

        Qwen3-8B & Qwen3-4B-Base
            & $17.82\!\rightarrow\!18.43$ ($+0.61$)
            & $18.43\!\rightarrow\!17.40$ ($-1.03$)
            & 0.0533 & $-1.680$ & -- \\

        Qwen3-14B & Qwen3-4B-Base
            & $18.43\!\rightarrow\!19.54$ ($+1.72$)
            & $18.43\!\rightarrow\!18.45$ ($+0.02$)
            & 0.0550 & $0.016$ & -- \\

        Qwen3-4B-SFT & Qwen3-4B-Base
            & $17.82\!\rightarrow\!36.03$ ($+18.21$)
            & $18.43\!\rightarrow\!36.62$ ($+18.19$)
            & 0.0708 & 0.999 & -- \\

        Qwen3-4B-Early-RL & Qwen3-4B-Base
            & $17.82\!\rightarrow\!39.91$ ($+22.09$)
            & $18.43\!\rightarrow\!40.85$ ($+22.42$)
            & 0.0465 & 1.015 & -- \\

        Qwen3-4B-RL & Qwen3-4B-Base
            & $17.82\!\rightarrow\!44.71$ ($+26.89$)
            & $18.43\!\rightarrow\!45.27$ ($+26.84$)
            & 0.0484 & 0.998 & 0.839 \\

        Qwen3-4B-Late-RL & Qwen3-4B-Base
            & $17.82\!\rightarrow\!50.10$ ($+32.28$)
            & $18.43\!\rightarrow\!46.61$ ($+28.18$)
            & 0.0719 & 0.873 & -- \\

        \midrule

        \multicolumn{7}{l}{
        \textit{Varying the reference; expert = Qwen3-4B-RL}}\\
        \addlinespace[2pt]

        Qwen3-4B-RL & Qwen3-4B-Early-RL
            & $39.91\!\rightarrow\!44.71$ ($+4.80$)
            & $18.43\!\rightarrow\!28.05$ ($+9.62$)
            & 0.1085 & 2.004 & 0.916 \\

        Qwen3-4B-RL & Qwen3-4B-Base
            & $17.82\!\rightarrow\!44.71$ ($+26.89$)
            & $18.43\!\rightarrow\!45.27$ ($+26.84$)
            & 0.0484 & 0.998 & 0.839 \\
        \rowcolor{gray!12}
        Qwen3-4B-RL & \shortstack{Qwen3-8B-Base}
            & $18.43\!\rightarrow\!44.71$ ($+26.28$)
            & $18.43\!\rightarrow\!42.30$ ($+23.87$)
            & 0.0000 & 0.908 & -- \\

        Qwen3-4B-RL & Qwen3-1.7B-Base
            & $8.92\!\rightarrow\!44.71$ ($+35.79$)
            & $18.43\!\rightarrow\!44.28$ ($+25.85$)
            & 0.1313 & 0.722 & -- \\

        Qwen3-4B-RL & Qwen2.5-3B
            & $1.04\!\rightarrow\!44.71$ ($+43.67$)
            & $18.43\!\rightarrow\!47.52$ ($+29.09$)
            & 1.2307 & 0.666 & 0.625 \\

        \midrule

        \multicolumn{7}{l}{
        \textit{Two RL increments with similar proxy gains}}\\
        \addlinespace[2pt]

        Qwen3-4B-RL & Qwen3-4B-Early-RL
            & $39.91\!\rightarrow\!44.71$ ($+4.80$)
            & $18.43\!\rightarrow\!28.05$ ($+9.62$)
            & 0.1085 & 2.004 & 0.916 \\

        Qwen3-4B-Late-RL & Qwen3-4B-RL
            & $44.71\!\rightarrow\!50.10$ ($+5.39$)
            & $18.43\!\rightarrow\!43.46$ ($+25.03$)
            & 0.1941 & 4.643 & 0.869 \\

        \bottomrule
    \end{tabular}
\end{table}

%% file: tables/rq3.tex
\begin{table}[h]
\centering
\scriptsize
\caption{
Transfer results across proxy pairs and targets.
The \colorbox{gray!10}{shaded} rows denote Qwen2.5-generation targets.
}
\label{tab:rq3_main}
\begin{tabularx}{\linewidth}{
    >{\raggedright\arraybackslash}X
    >{\raggedright\arraybackslash}X
    >{\raggedright\arraybackslash}X
    cccc
}
\toprule
Proxy Expert & Proxy Reference & Target
& $K_\text{Base}$ & $G_T$ & $R_{\mathrm{perf}}$ & Avg \\
\midrule

Qwen3-1.7B-RL & Qwen3-1.7B & Qwen3-1.7B
& 0.000 & 16.11 & \phantom{-}0.993 & 25.03 \\
Qwen3-1.7B-RL & Qwen3-1.7B & Qwen3-4B
& 0.106 & 13.65 & \phantom{-}0.841 & 31.47 \\
Qwen3-1.7B-RL & Qwen3-1.7B & Qwen3-8B
& 0.111 & 9.78 & \phantom{-}0.603 & 28.21 \\
\rowcolor{gray!12}
Qwen3-1.7B-RL & Qwen3-1.7B & Qwen2.5-7B
& 0.462 & -1.64 & -0.101 & \phantom{0}0.15 \\
\rowcolor{gray!12}
Qwen3-1.7B-RL & Qwen3-1.7B & Qwen2.5-7B-Ins
& 0.656 & -1.69 & -0.104 & \phantom{0}5.17 \\
\rowcolor{gray!12}
Qwen3-1.7B-RL & Qwen3-1.7B & Qwen2.5-7B-Math
& 0.628 & -0.08 & -0.005 & \phantom{0}0.69 \\

\midrule

Qwen3-4B-RL & Qwen3-4B & Qwen3-1.7B
& 0.123 & 17.55 & \phantom{-}0.653 & 26.47 \\
Qwen3-4B-RL & Qwen3-4B & Qwen3-4B
& 0.000 & 26.88 & \phantom{-}1.000 & 44.70 \\
Qwen3-4B-RL & Qwen3-4B & Qwen3-8B
& 0.048 & 26.84 & \phantom{-}0.998 & 45.27 \\
\rowcolor{gray!12}
Qwen3-4B-RL & Qwen3-4B & Qwen2.5-7B
& 0.342 & -1.74 & -0.065 & \phantom{0}0.05 \\
\rowcolor{gray!12}
Qwen3-4B-RL & Qwen3-4B & Qwen2.5-7B-Ins
& 0.594 & -0.05 & -0.002 & \phantom{0}6.81 \\
\rowcolor{gray!12}
Qwen3-4B-RL & Qwen3-4B & Qwen2.5-7B-Math
& 0.550 & 1.44 & \phantom{-}0.054 & \phantom{0}2.21 \\

\midrule

Qwen3-4B-RL & Qwen2.5-3B & Qwen3-1.7B
& 1.050 & 17.54 & \phantom{-}0.402 & 26.46 \\
Qwen3-4B-RL & Qwen2.5-3B & Qwen3-4B
& 1.299 & 28.14 & \phantom{-}0.644 & 45.96 \\
Qwen3-4B-RL & Qwen2.5-3B & Qwen3-8B
& 1.231 & 29.09 & \phantom{-}0.666 & 47.52 \\
\rowcolor{gray!12}
Qwen3-4B-RL & Qwen2.5-3B & Qwen2.5-7B
& 0.099 & 6.74 & \phantom{-}0.154 & \phantom{0}8.53 \\
\rowcolor{gray!12}
Qwen3-4B-RL & Qwen2.5-3B & Qwen2.5-7B-Ins
& 0.162 & 2.25 & \phantom{-}0.052 & \phantom{0}9.11 \\
\rowcolor{gray!12}
Qwen3-4B-RL & Qwen2.5-3B & Qwen2.5-7B-Math
& 0.151 & 5.76 & \phantom{-}0.132 & \phantom{0}6.53 \\

\midrule

Qwen3-4B-RL & Qwen3-1.7B & Qwen3-1.7B
& 0.000 & 18.67 & \phantom{-}0.522 & 27.59 \\
Qwen3-4B-RL & Qwen3-1.7B & Qwen3-4B
& 0.124 & 27.87 & \phantom{-}0.779 & 45.69 \\
Qwen3-4B-RL & Qwen3-1.7B & Qwen3-8B
& 0.131 & 25.85 & \phantom{-}0.722 & 44.28 \\
\rowcolor{gray!12}
Qwen3-4B-RL & Qwen3-1.7B & Qwen2.5-7B
& 0.374 & -1.79 & -0.050 & \phantom{0}0.00 \\
\rowcolor{gray!12}
Qwen3-4B-RL & Qwen3-1.7B & Qwen2.5-7B-Ins
& 0.561 & -2.46 & -0.069 & \phantom{0}4.40 \\
\rowcolor{gray!12}
Qwen3-4B-RL & Qwen3-1.7B & Qwen2.5-7B-Math
& 0.562 & 2.72 & \phantom{-}0.076 & \phantom{0}3.49 \\

\bottomrule
\end{tabularx}
\end{table}

%% file: tables/rq4.tex
\begin{table*}[h]
\centering
\scriptsize
\setlength{\tabcolsep}{5pt}
\caption{Representative successful and failed settings performance on both math and
code.}
\label{tab:cross_domain}
\begin{tabular}{l l l cc cc}
\toprule
& & & \multicolumn{2}{c}{Math} & \multicolumn{2}{c}{Code} \\
\cmidrule(lr){4-5}\cmidrule(lr){6-7}
Proxy Expert & Proxy Reference & Target & Avg & $G_T$ & Avg & $G_T$ \\
\midrule
\multicolumn{7}{l}{\textit{Same-generation transfer (Qwen3 $\rightarrow$ Qwen3)}}\\
Qwen3-4B-RL & Qwen3-4B   & Qwen3-8B & 45.27 & $+26.8$ & 56.44 & $+2.3$ \\
Qwen3-4B-RL & Qwen3-1.7B & Qwen3-8B & 44.28 & $+25.9$ & 56.03 & $+1.9$ \\
\midrule
\multicolumn{7}{l}{\textit{Cross-generation transfer (Qwen3 proxy $\rightarrow$ Qwen2.5 target)}}\\
Qwen3-4B-RL & Qwen3-4B & Qwen2.5-7B & \phantom{0}0.05 & $-1.7$ & 22.83 & $-6.4$ \\
\midrule
\multicolumn{7}{l}{\textit{Weak / reversed source}}\\
Qwen3-4B & Qwen3-1.7B & Qwen3-8B & 21.85 & $+3.4$ & 51.53 & $-2.6$ \\
Qwen3-8B & Qwen3-4B   & Qwen3-8B & 17.41 & $-1.0$ & 48.21 & $-6.0$ \\
\midrule
\multicolumn{7}{l}{\textit{Aligned cross-generation expert (Qwen2.5-based proxy $\rightarrow$ Qwen3-8B)}}\\
DeepSeek-R1-Distill & Qwen2.5-Math/Coder & Qwen3-8B & 43.06 & $+24.6$ & 63.27 & $+9.1$ \\
\bottomrule
\end{tabular}
\end{table*}

%% file: sections/conclusion.tex
\section{Conclusion}
\label{sec:conclusion}
\vspace{-1em}
In this work, we revisited relative policy difference transfer through the
lens of implicit reward. This perspective leads to A2-OPD, an any-to-any
formulation that decouples the proxy pair defining the relative difference from the target receiving it, with conventional OPD recovered as a special case. Our experiments further characterize the opportunities and limits of this generality. Policy differences induced by task-directed training, including RL and SFT, transfer consistently, whereas differences due to model scaling alone are often ineffective. We also find that the proxy expert supplies the capability to be transferred, while the proxy reference determines the residual presented to the target. Finally, transfer depends on target compatibility: a useful difference must induce an update that the target can follow and generalize from, and an appropriate reference can recover some otherwise unsuccessful transfers. Together, these results clarify when any-to-any relative policy difference transfer provides effective supervision and offer practical guidance for selecting proxy pairs and targets.

%% file: sections/appendix.tex
\newpage
\appendix
\renewcommand{\thesection}{\Alph{section}}
\setcounter{section}{0}

\noindent{\LARGE\textbf{Appendix}\par}\normalsize

\vspace{10pt}
{\large \textbf{Contents}}
\startcontents[appendices]
\printcontents[appendices]{l}{1}{\setcounter{tocdepth}{3}}

\section{Complete Theoretical Derivation}
\label{app:implicit_reward_derivation}

\subsection{Implicit Reward Encoded by the Proxy Pair}
\label{app:proxy_implicit_reward}

Fix a state $s$. The state-wise optimization problem in
Eq.~\ref{eq:proxy_reward_optimization} is
\begin{equation}
\max_{\pi(\cdot\mid s)}
\sum_a \pi(a\mid s)
\left[
r^{*}(s,a)
-
\tau\log\frac{\pi(a\mid s)}{\pi_\phi(a\mid s)}
\right]
\quad
\text{subject to}
\quad
\sum_a\pi(a\mid s)=1.
\label{eq:proxy_statewise_objective}
\end{equation}
Introducing a Lagrange multiplier $\eta_\phi(s)$, the stationarity
condition is
\begin{equation}
r^{*}(s,a)
-
\tau\left(
\log\frac{\pi(a\mid s)}{\pi_\phi(a\mid s)}+1
\right)
+
\eta_\phi(s)
=0.
\label{eq:proxy_stationarity}
\end{equation}
Solving this condition and imposing normalization gives
\begin{equation}
\pi_\phi^{+}(a\mid s)
=
\frac{1}{Z_\phi(s)}
\pi_\phi(a\mid s)
\exp\left(
\frac{r^{*}(s,a)}{\tau}
\right),
\qquad
Z_\phi(s)
=
\sum_{a'}
\pi_\phi(a'\mid s)
\exp\left(
\frac{r^{*}(s,a')}{\tau}
\right).
\label{eq:proxy_induced_policy}
\end{equation}
Therefore, on the support of $\pi_\phi$,
\begin{equation}
\Delta_\phi(s,a)
=
\log\frac{\pi_\phi^{+}(a\mid s)}{\pi_\phi(a\mid s)}
=
\frac{1}{\tau}r^{*}(s,a)
-
\log Z_\phi(s),
\end{equation}
which establishes Eq.~\ref{eq:proxy_implicit_reward}.

\subsection{Closed-Form Policy Induced by the Any-to-Any Objective}
\label{app:any_to_any_induced_policy}

Fix a state $s$. The state-wise objective in
Eq.~\ref{eq:any_to_any_objective} is
\begin{equation}
\mathcal{J}_s(\pi)
=
\sum_a \pi(a\mid s)
\left[
\alpha\Delta_\phi(s,a)
-
\beta\log
\frac{\pi(a\mid s)}{\pi_{\mathrm{ref}}(a\mid s)}
\right].
\label{eq:any_to_any_statewise_objective}
\end{equation}

Introducing a Lagrange multiplier $\eta(s)$ for the constraint
$\sum_a\pi(a\mid s)=1$, the first-order optimality condition is
\begin{equation}
\alpha\Delta_\phi(s,a)
-
\beta\left(
\log\frac{\pi(a\mid s)}{\pi_{\mathrm{ref}}(a\mid s)}+1
\right)
+
\eta(s)
=0.
\label{eq:any_to_any_stationarity}
\end{equation}

Solving Eq.~\ref{eq:any_to_any_stationarity} and enforcing normalization
yields

\begin{equation}
\pi_\theta^{*}(a\mid s)
=
\frac{1}{Z_\theta(s)}
\pi_{\mathrm{ref}}(a\mid s)
\exp\left(
\frac{\alpha}{\beta}\Delta_\phi(s,a)
\right),
\label{eq:any_to_any_induced_policy}
\end{equation}

where
\begin{equation}
Z_\theta(s)
=
\sum_{a'}
\pi_{\mathrm{ref}}(a'\mid s)
\exp\left(
\frac{\alpha}{\beta}\Delta_\phi(s,a')
\right).
\label{eq:any_to_any_normalizer}
\end{equation}

Equivalently, using Eq.~\ref{eq:relative_policy_difference},

\begin{equation}
\pi_\theta^{*}(a\mid s)
\propto
\pi_{\mathrm{ref}}(a\mid s)
\left(
\frac{\pi_\phi^{+}(a\mid s)}{\pi_\phi(a\mid s)}
\right)^{\alpha/\beta}.
\label{eq:any_to_any_ratio_form}
\end{equation}

Thus, the proxy pair determines the relative reweighting, whereas
$\pi_{\mathrm{ref}}$ determines the distribution to which that reweighting
is applied. For $\beta>0$, Eq.~\ref{eq:any_to_any_statewise_objective} is
strictly concave in $\pi(\cdot\mid s)$ on the support of $\pi_{\mathrm{ref}}$, so Eq.~\ref{eq:any_to_any_induced_policy} is the unique
state-wise maximizer on that support.

\subsection{Gradient of the KL-Regularized Objective}
\label{app:any_to_any_policy_gradient}

We derive the conditional gradient used by the on-policy surrogate in
Eq.~\ref{eq:A2_policy_gradient}. During each update, the states collected by
the current rollout policy are treated as fixed samples; equivalently, no
gradient is propagated through $d_{\pi_\theta}$. Using
\begin{equation}
\ell_\theta(s,a)
=
\log\frac{\pi_\theta(a\mid s)}{\pi_{\mathrm{ref}}(a\mid s)},
\qquad
u_\theta(s,a)
=
\alpha\Delta_\phi(s,a)-\beta\ell_\theta(s,a),
\label{eq:appendix_target_score_definitions}
\end{equation}
in agreement with Eq.~\ref{eq:A2_policy_gradient}, the state-wise objective
can be written as
\begin{equation}
\mathcal{J}_s(\theta)
=
\sum_a
\pi_\theta(a\mid s)u_\theta(s,a).
\label{eq:any_to_any_statewise_theta_objective}
\end{equation}
Although $\Delta_\phi$ is fixed, $u_\theta$ depends on $\theta$ through
$\ell_\theta$. Differentiating Eq.~\ref{eq:any_to_any_statewise_theta_objective}
gives

\begin{align}
\nabla_\theta\mathcal{J}_s(\theta)
&=
\sum_a
\nabla_\theta\pi_\theta(a\mid s)u_\theta(s,a)
-
\beta\sum_a
\pi_\theta(a\mid s)\nabla_\theta\ell_\theta(s,a).
\label{eq:any_to_any_gradient_expansion}
\end{align}

Because $\pi_{\mathrm{ref}}$ is fixed,
$\nabla_\theta\ell_\theta(s,a)
=\nabla_\theta\log\pi_\theta(a\mid s)$. Consequently, the second term in
Eq.~\ref{eq:any_to_any_gradient_expansion} vanishes:
\begin{align}
-\beta\sum_a
\pi_\theta(a\mid s)\nabla_\theta\ell_\theta(s,a)
&=
-\beta\sum_a\nabla_\theta\pi_\theta(a\mid s)
\nonumber\\
&=
-\beta\nabla_\theta\sum_a\pi_\theta(a\mid s)
=0.
\label{eq:any_to_any_explicit_gradient_zero}
\end{align}
Applying
$\nabla_\theta\pi_\theta
=\pi_\theta\nabla_\theta\log\pi_\theta$
to the remaining term yields
\begin{equation}
\nabla_\theta\mathcal{J}_s(\theta)
=
\mathbb{E}_{a\sim\pi_\theta(\cdot\mid s)}
\left[
u_\theta(s,a)
\nabla_\theta\log\pi_\theta(a\mid s)
\right].
\label{eq:any_to_any_statewise_policy_gradient}
\end{equation}
Taking the empirical expectation over states sampled from
$d_{\pi_\theta}$ yields the gradient estimator in
Eq.~\ref{eq:A2_policy_gradient}.

If Eq.~\ref{eq:any_to_any_objective} is instead interpreted as a full
trajectory objective and gradients through the policy-induced state
distribution are retained, the exact score-function weight is the
return-to-go (or action value) induced by $u_\theta$, rather than the
single-step quantity $u_\theta(s,a)$. Eq.~\ref{eq:A2_policy_gradient}
is therefore the exact conditional gradient of the state-wise on-policy
surrogate used here.

\section{Experimental and Implementation Details}
\label{app:experiment_setup}

\paragraph{Models.}
For math, the regular RL expert is trained for 500 steps, while the early
and late experts are taken at 100 and 1,200 steps, respectively. For code,
the RL expert is trained for 300 steps. Qwen3-4B-SFT denotes Qwen3-4B
fine-tuned on correct trajectories rejection-sampled from DeepMath-103K using
Skywork-7B: 14,716 of the 15,491 trajectories are used for training and
775 for validation. SFT is configured for three epochs (229 steps) and
stops at step 70 under early stopping with a patience of 5.

\paragraph{Training Details.}
Our implementation follows the on-policy training setup of P-OPD. For each prompt, the target policy samples eight responses with temperature $1.0$ and top-$p$ $1.0$. We use a training batch size of $1024$, a learning rate of $1\times10^{-6}$, and a maximum response length of $16{,}384$ tokens. The proxy expert, proxy reference, and target reference remain frozen, and gradients are applied only to the target policy. For models with different tokenizers, we use text as an intermediate representation to map target tokens into the proxy token space and retain only successfully aligned tokens for training. We train for $80$ steps in the math domain and $150$ steps in the code domain. In both domains, the gradient norm has reached a stable plateau by the end of training. Math experiments use DeepMath-103K, while code experiments use Eurus-RL-Code. Unless
otherwise stated, math results are reported as Mean@16 and code results as Mean@8.

\paragraph{Metrics.}
\label{app:metrics}
Let $P(\pi)$ denote the average task performance of policy $\pi$ over
the evaluation benchmarks. We define the performance gain represented
by a proxy pair and the gain realized by the target as
\begin{equation}
G_P = P(\pi_\phi^+) - P(\pi_\phi),
\qquad
G_T = P(\pi_\theta^*) - P(\pi_{\mathrm{ref}}),
\end{equation}
where $\pi_\theta^*$ denotes the optimized target policy.
Their ratio,
\begin{equation}
R_{\mathrm{perf}} = \frac{G_T}{G_P},
\end{equation}
measures how much of the proxy performance gain is realized by the
target.

We additionally report distributional metrics on a fixed set of 500
prompts randomly sampled from the DeepMath-103K training set. For each
configuration, we generate responses using its optimized target policy and
all policies teacher-forcing in that configuration on the resulting trajectories. Distributional metrics are then computed over the same
token-level states within each configuration. Let
\begin{equation}
\mathcal{K}(p,q)
=
\mathbb{E}_{s}
\left[
D_{\mathrm{KL}}
\bigl(p(\cdot\mid s)\,\|\,q(\cdot\mid s)\bigr)
\right]
\end{equation}
denote the forward KL divergence averaged over non-padding response
tokens. The proxy-policy difference is measured by
\begin{equation}
K_\text{Proxy} = \mathcal{K}(\pi_\phi^+,\pi_\phi),
\end{equation}
and the initial proxy--target mismatch by
\begin{equation}
K_\text{Base} = \mathcal{K}(\pi_{\mathrm{ref}},\pi_\phi).
\end{equation}
Here, $K_B$ measures only the initial distributional compatibility
between the proxy reference and target reference; it does not depend on
the proxy expert.

To characterize the distribution implied by the proxy difference, we
define
\begin{equation}
\pi_\text{Induced}(a\mid s)
=
\frac{
\pi_{\mathrm{ref}}(a\mid s)
\exp\!\left(\frac{\alpha}{\beta}\Delta_\phi(s,a)\right)
}{
\sum_{a'}
\pi_{\mathrm{ref}}(a'\mid s)
\exp\!\left(\frac{\alpha}{\beta}\Delta_\phi(s,a')\right)
},
\end{equation}
where
$\Delta_\phi(s,a)=
\log\pi_\phi^+(a\mid s)-\log\pi_\phi(a\mid s)$.
We then define the induced and realized target-policy changes as
\begin{equation}
K_\text{Induced} = \mathcal{K}(\pi_\text{Induced},\pi_{\mathrm{ref}}),
\qquad
K_\text{Target} = \mathcal{K}(\pi_\theta^*,\pi_{\mathrm{ref}}),
\end{equation}
respectively. Their residual discrepancy is
\begin{equation}
K_{\mathrm{gap}} = \mathcal{K}(\pi_\theta^*,\pi_\text{Induced}),
\end{equation}
and the distributional realization ratio is
\begin{equation}
R_{\mathrm{KL}} = \frac{K_\text{Target}}{K_\text{Induced}}.
\end{equation}
While $R_{\mathrm{KL}}$ compares the magnitudes of the induced and
realized policy changes, $K_{\mathrm{gap}}$ directly measures the
remaining difference between their distributions.

Finally, we report the performance gain per unit of policy change:
\begin{equation}
E_P = \frac{G_P}{K_\text{Proxy}},
\qquad
E_T = \frac{G_T}{K_\text{Target}}.
\end{equation}
These quantities measure the distributional efficiency of the proxy
update and target update, respectively. For the overlap analysis, we
compute the top-$k$ token overlap between policies $p$ and $q$ as
\begin{equation}
O_k(p,q)
=
\mathbb{E}_{s}
\left[
\frac{
\left|
\operatorname{TopK}(p(\cdot\mid s))
\cap
\operatorname{TopK}(q(\cdot\mid s))
\right|
}{k}
\right],
\end{equation}
using the same target (Qwen3-8B) generated states for both policies.

\section{Additional Experimental Results}
\section{Complete Experimental Results and Analyses}

We report the complete validation performance of all base and expert checkpoints used in our experiments, followed by the full validation results and distributional evaluations for each experimental configuration presented in the main text.

\input{tables/base_expert-performance}

\input{tables/rq1_3tables_version}

\input{tables/rq2-1_3tables_version}

\input{tables/rq2-2_3tables_version}

\input{tables/rq3_3tables_version}

%% file: tables/base_expert-performance.tex
\begin{table*}[t]
\centering
\scriptsize
\setlength{\tabcolsep}{4pt}
\caption{Base model and proxy expert performance.}
\label{app:base_expert}
\begin{subtable}{\textwidth}
\centering
\caption{Math (averaged over AIME24/25/26, SMT25, CMIMC25, HMMT25-Feb/Nov, HMMT26-Feb).}

\resizebox{\linewidth}{!}{%
\begin{tabular}{@{}l*{10}{c}@{}}
\toprule
Model & AIME24 & AIME25 & AIME26 & SMT25 & CMIMC25
& HMMT25(F) & HMMT25(N) & HMMT26(F) & Avg & Gain \\
\midrule
Qwen3-1.7B                    & 11.88 & 11.04 &  9.17 & 16.39 &  5.00 &  5.83 &  5.21 &  6.82 &  8.92 & -- \\
Qwen3-4B                      & 24.79 & 21.67 & 16.25 & 28.18 & 12.34 & 11.67 &  8.75 & 13.26 & 17.82 & -- \\
Qwen3-8B                      & 25.21 & 21.04 & 16.25 & 27.71 & 13.75 & 10.62 & 11.88 & 15.72 & 18.43 & -- \\
Qwen2.5-3B                    &  1.40 &  0.42 &  0.62 &  2.48 &  0.62 &  0.00 &  1.25 &  0.57 &  1.04 & -- \\
Qwen2.5-7B                    &  3.54 &  1.25 &  1.67 &  3.42 &  0.47 &  0.21 &  2.08 &  0.95 &  1.79 & -- \\
Qwen2.5-7B-Instruct           &  9.38 &  7.71 &  5.42 & 13.56 &  4.53 &  2.08 &  4.79 &  3.41 &  6.86 & -- \\
Qwen2.5-Math-7B               &  1.88 &  0.83 &  0.21 &  1.42 &  0.31 &  0.21 &  0.62 &  0.38 &  0.77 & $-1.02$ \\
\midrule
Qwen3-4B-SFT-Math             & 49.38 & 42.50 & 45.42 & 45.64 & 21.88 & 24.17 & 32.92 & 24.81 & 36.03 & +18.21 \\
Qwen3-1.7B Math-RL (step 500) & 35.62 & 31.67 & 29.58 & 34.79 & 16.72 & 19.58 & 15.21 & 17.99 & 25.15 & +16.23 \\
Qwen3-4B Math-RL (step 100)   & 55.00 & 49.58 & 49.58 & 47.88 & 27.66 & 27.29 & 32.92 & 29.36 & 39.91 & +22.09 \\
Qwen3-4B Math-RL (step 500)   & 58.75 & 55.83 & 56.88 & 50.83 & 30.63 & 33.12 & 38.33 & 33.33 & 44.71 & +26.89 \\
Qwen3-4B Math-RL (step 1200)  & 69.38 & 60.83 & 62.50 & 55.07 & 32.81 & 38.12 & 44.58 & 37.50 & 50.10 & +32.28 \\
DeepSeek-R1-Distill-Qwen-7B   & 46.46 & 37.92 & 43.12 & 44.46 & 21.09 & 21.67 & 25.00 & 26.89 & 33.74 & +32.97 \\
Skywork-OR1-Math-7B           & 63.54 & 48.54 & 31.09 & 57.29 & 30.83 & 37.71 & 51.18 & 31.25 & 43.93 & +42.14 \\
\bottomrule
\end{tabular}
}
\end{subtable}
\vspace{1em}

\begin{subtable}{\textwidth}
\centering
\caption{Code (averaged over HumanEval+, MBPP+, LiveCodeBench).}
\begin{tabular}{l ccc c c}
\toprule
Model & HumanEval+ & MBPP+ & LCB & Avg & Gain \\
\midrule
Qwen2.5-7B                  & 42.53 & 41.04 &  4.14 & 29.24 & -- \\
Qwen3-1.7B                  & 61.10 & 52.10 & 11.70 & 41.65 & -- \\
Qwen3-4B                    & 75.60 & 63.00 & 18.00 & 52.20 & -- \\
Qwen3-8B                    & 79.70 & 66.50 & 16.30 & 54.16 & -- \\
Qwen2.5-Coder-7B            & 33.77 & 36.14 &  0.93 & 23.61 & $-5.63$ \\
\midrule
DeepSeek-R1-Distill-Qwen-7B & 69.36 & 56.91 & 21.93 & 49.40 & +25.79 \\
Qwen3-4B Code-RL (step 300) & 77.70 & 62.00 & 19.30 & 52.99 & +0.79 \\
\bottomrule
\end{tabular}
\end{subtable}
\end{table*}

%% file: tables/rq1_3tables_version.tex
\begin{table}[htbp]
  \centering
  \scriptsize
  \setlength{\tabcolsep}{3.5pt}
  \begin{subtable}{\textwidth}
    \centering
    \caption{Model configurations and their identifiers.}
    \label{tab:rq1-configurations}
    \begin{tabular*}{\linewidth}{@{\extracolsep{\fill}}clll@{}}
      \toprule
      ID & Proxy expert & Proxy reference & Target \\
      \midrule
      1  & Qwen3-1.7B-RL                  & Qwen3-1.7B   & Qwen3-8B \\
      2  & Qwen3-1.7B-RL                  & Qwen3-1.7B   & Qwen3-4B \\
      3  & Qwen3-4B-RL                    & Qwen3-8B     & Qwen3-8B \\
      4  & Qwen3-4B-RL                    & Qwen3-4B     & Qwen3-8B \\
      5  & Qwen3-4B-SFT                   & Qwen3-4B     & Qwen3-8B \\
      6  & Qwen3-4B                       & Qwen3-1.7B        & Qwen3-8B \\
      7  & Qwen3-8B                       & Qwen3-4B          & Qwen3-8B \\
      8  & Qwen3-14B                      & Qwen3-4B          & Qwen3-8B \\
      9  & Qwen3-14B                      & Qwen3-1.7B          & Qwen3-8B \\
      10  & Qwen3-8B                       & Qwen2.5-7B        & Qwen3-8B \\
      11 & Skywork-7B                     & Qwen2.5-7B        & Qwen3-8B \\
      12 & Qwen3-4B-RL                    & Qwen3-1.7B   & Qwen3-8B \\
      13 & DeepSeek-R1-Distill-Qwen-7B    & Qwen2.5-Math-7B   & Qwen3-8B \\
      \bottomrule
    \end{tabular*}
  \end{subtable}

  \vspace{0.75em}

  \begin{subtable}{\textwidth}
    \centering
    \caption{Validation results for each configuration.}
    \label{tab:rq1-validation}
    \begin{tabular*}{\linewidth}{@{\extracolsep{\fill}}c*{9}{c}@{}}
        \toprule
        ID & AIME24 & AIME25 & AIME26 & SMT25 & CMIMC25 & HMMT25 (Feb) & HMMT25 (Nov) & HMMT26 (Feb) & Avg \\
        \midrule
        1  & 40.625 & 33.750 & 31.667 & 37.146 & 17.656 & 14.583 & 25.625 & 24.621 & 28.209 \\
        2  & 45.208 & 39.583 & 36.458 & 40.684 & 20.469 & 20.000 & 26.042 & 23.295 & 31.467 \\
        3  & 57.083 & 49.583 & 55.208 & 51.297 & 27.188 & 29.167 & 36.667 & 32.197 & 42.299 \\
        4  & 63.750 & 52.500 & 57.083 & 52.476 & 30.156 & 31.458 & 40.833 & 33.902 & 45.270 \\
        5  & 53.750 & 42.292 & 44.792 & 44.575 & 23.438 & 23.542 & 34.583 & 25.947 & 36.615 \\
        6  & 32.917 & 26.875 & 22.500 & 30.307 & 14.688 & 11.042 & 17.917 & 18.561 & 21.851 \\
        7  & 26.458 & 23.125 & 16.250 & 27.123 & 12.500 & 9.792  & 7.708  & 16.288 & 17.405 \\
        8  & 26.667 & 22.083 & 18.542 & 28.420 & 15.000 & 10.833 & 9.375  & 16.667 & 18.448 \\

        10  & 40.625 & 29.375 & 27.917 & 35.495 & 20.156 & 15.833 & 20.000 & 23.864 & 26.658 \\
        11 & 70.000 & 59.583 & 61.667 & 55.071 & 37.969 & 36.042 & 46.875 & 38.068 & 50.659 \\
        12 & 62.917 & 49.792 & 56.250 & 51.061 & 30.938 & 31.458 & 40.417 & 31.439 & 44.284 \\
        13 & 63.333 & 46.042 & 54.792 & 50.000 & 30.313 & 27.292 & 39.375 & 33.333 & 43.060 \\
        \bottomrule
      \end{tabular*}
  \end{subtable}

  \vspace{0.75em}

  \begin{subtable}{\textwidth}
    \centering
    \caption{Distributional and efficiency metrics for each configuration.}
    \label{tab:rq1-metrics}
    \begin{tabular*}{\linewidth}{@{\extracolsep{\fill}}c*{11}{c}@{}}
      \toprule
      ID & $G_P$ & $G_T$ & $K_P$ & $K_B$ & $K_I$ & $K_T$ & $K_{\mathrm{gap}}$ & $E_P$ & $R_{\mathrm{KL}}$ & $R_{\mathrm{perf}}$ & $E_T$ \\
      \midrule
      1  & 16.23 & 9.779  & 0.0593 & 0.1110 & 0.0773 & 0.0587 & 0.0157 & 273.6931    & 0.7594      & 0.6025       & 166.5928 \\
      2  & 16.23 & 13.647 & 0.0631 & 0.1055 & 0.0766 & 0.0595 & 0.0148 & 257.2108    & 0.7768      & 0.8409       & 229.3613 \\
      3$^{\dagger}$  & 26.89 & 23.869 & --     & --     & --     & --     & --     & --          & --          & --           & -- \\
      4  & 26.89 & 26.840 & 0.0966 & 0.0484 & 0.1066 & 0.0978 & 0.0135 & 278.3644    & 0.9174      & 0.9981       & 274.4376 \\
      5  & 18.21 & 18.185 & 0.1401 & 0.0708 & 0.1725 & 0.1275 & 0.0269 & 129.9897    & 0.7389      & 0.9986       & 142.6621 \\
      6  & 8.90  & 3.421  & 0.1390 & 0.1320 & 0.4181 & 0.0474 & 0.2188 & 64.0288     & 0.1134      & 0.3844       & 72.1730 \\
      7  & 0.61  & -1.025 & 0.0593 & 0.0533 & 0.1379 & 0.0239 & 0.0776 & 10.2867     & 0.1733      & -1.6803      & -42.8870 \\
      8  & 1.72  & 0.018  & 0.0896 & 0.0550 & 0.1730 & 0.0202 & 0.1023 & 19.2005 & 0.1166 & 0.01047 & 0.8928 \\
      9  & 16.64 & 8.228  & 0.3363 & 1.0379 & 0.1522 & 0.0337 & 0.1155 & 49.4796     & 0.2214      & 0.4945       & 244.1543 \\
      10 & 42.14 & 32.229 & 0.4417 & 1.0604 & 0.2292 & 0.0935 & 0.2297 & 95.4041     & 0.4079      & 0.7648       & 344.6952 \\
      11 & 35.79 & 25.854 & 0.2384 & 0.1313 & 0.3531 & 0.1321 & 0.1748 & 150.1258    & 0.3741      & 0.7224       & 195.7154 \\
      12 & 32.97 & 24.630 & 0.5915 & 2.8489 & 0.1785 & 0.1024 & 0.1420 & 55.7396     & 0.5737      & 0.7470       & 240.5273 \\
      \bottomrule
      \end{tabular*}
    \raggedright
    $^{\dagger}$Distributional and efficiency metrics are not reported for ID~3 because this configuration uses standard OPD.
  \end{subtable}
  \caption{Configurations, validation results, and distributional and efficiency metrics for policy difference experiments.}
  \label{tab:rq1_full}
\end{table}

%% file: tables/rq2-1_3tables_version.tex
\begin{table}[t]
  \centering
  \scriptsize
  \setlength{\tabcolsep}{3.5pt}
  \begin{subtable}{\textwidth}
    \centering
    \caption{Model configurations and their identifiers.}
    \label{tab:rq2-1-configurations}
     \begin{tabular*}{\linewidth}{@{\extracolsep{\fill}}clll@{}}
      \toprule
      ID & Proxy expert & Proxy reference & Target \\
      \midrule
      1  & Qwen3-1.7B-RL                  & Qwen3-1.7B   & Qwen3-8B \\
      2  & Qwen3-1.7B-RL                  & Qwen3-1.7B   & Qwen3-4B \\
      3  & Qwen3-4B-RL                    & Qwen3-8B     & Qwen3-8B \\
      4  & Qwen3-4B-RL                    & Qwen3-4B     & Qwen3-8B \\
      5  & Qwen3-4B-SFT                   & Qwen3-4B     & Qwen3-8B \\
      6  & Qwen3-4B                       & Qwen3-1.7B        & Qwen3-8B \\
      7  & Qwen3-8B                       & Qwen3-4B          & Qwen3-8B \\
      8  & Qwen3-14B                      & Qwen3-4B          & Qwen3-8B \\
      9  & Qwen3-14B                      & Qwen3-1.7B          & Qwen3-8B \\
      10  & Qwen3-8B                       & Qwen2.5-7B        & Qwen3-8B \\
      11 & Skywork-7B                     & Qwen2.5-7B        & Qwen3-8B \\
      12 & Qwen3-4B-RL                    & Qwen3-1.7B   & Qwen3-8B \\
      13 & DeepSeek-R1-Distill-Qwen-7B    & Qwen2.5-Math-7B   & Qwen3-8B \\
      \bottomrule
    \end{tabular*}
  \end{subtable}

  \vspace{0.75em}

  \begin{subtable}{\textwidth}
    \centering
    \caption{Validation results for each configuration.}
    \label{tab:rq2-1-validation}
    \begin{tabular*}{\linewidth}{@{\extracolsep{\fill}}c*{9}{c}@{}}
        \toprule
        ID & AIME24 & AIME25 & AIME26 & SMT25 & CMIMC25 & HMMT25 (Feb) & HMMT25 (Nov) & HMMT26 (Feb) & Avg \\
        \midrule
        1  & 40.625 & 33.750 & 31.667 & 37.146 & 17.656 & 14.583 & 25.625 & 24.621 & 28.209 \\
        2  & 45.208 & 39.583 & 36.458 & 40.684 & 20.469 & 20.000 & 26.042 & 23.295 & 31.467 \\
        3  & 57.083 & 49.583 & 55.208 & 51.297 & 27.188 & 29.167 & 36.667 & 32.197 & 42.299 \\
        4  & 63.750 & 52.500 & 57.083 & 52.476 & 30.156 & 31.458 & 40.833 & 33.902 & 45.270 \\
        5  & 53.750 & 42.292 & 44.792 & 44.575 & 23.438 & 23.542 & 34.583 & 25.947 & 36.615 \\
        6  & 32.917 & 26.875 & 22.500 & 30.307 & 14.688 & 11.042 & 17.917 & 18.561 & 21.851 \\
        7  & 26.458 & 23.125 & 16.250 & 27.123 & 12.500 & 9.792  & 7.708  & 16.288 & 17.405 \\
        8  & 26.667 & 22.083 & 18.542 & 28.420 & 15.000 & 10.833 & 9.375  & 16.667 & 18.448 \\
        9  & 30.625 & 24.375 & 23.333 & 30.542 & 15.938 & 13.333 & 16.250 & 18.182 & 21.572 \\
        10  & 40.625 & 29.375 & 27.917 & 35.495 & 20.156 & 15.833 & 20.000 & 23.864 & 26.658 \\
        11 & 70.000 & 59.583 & 61.667 & 55.071 & 37.969 & 36.042 & 46.875 & 38.068 & 50.659 \\
        12 & 62.917 & 49.792 & 56.250 & 51.061 & 30.938 & 31.458 & 40.417 & 31.439 & 44.284 \\
        13 & 63.333 & 46.042 & 54.792 & 50.000 & 30.313 & 27.292 & 39.375 & 33.333 & 43.060 \\
        \bottomrule
      \end{tabular*}
  \end{subtable}

  \vspace{0.75em}

  \begin{subtable}{\textwidth}
    \centering
    \caption{Distributional and efficiency metrics for each configuration.}
    \label{tab:rq2-1-metrics}
    \begin{tabular*}{\linewidth}{@{\extracolsep{\fill}}c*{12}{c}@{}}
      \toprule
      ID & $G_P$ & $G_T$ & $K_P$ & $K_B$ & $K_I$ & $K_T$ & $K_{\mathrm{gap}}$ & $E_P$ & $R_{\mathrm{KL}}$ & $R_{\mathrm{perf}}$ & $E_T$ \\
      \midrule
      1  & 16.23 & 9.779  & 0.0593 & 0.1110 & 0.0773 & 0.0587 & 0.0157 & 273.6931    & 0.7594      & 0.6025       & 166.5928 \\
      2  & 16.23 & 13.647 & 0.0631 & 0.1055 & 0.0766 & 0.0595 & 0.0148 & 257.2108    & 0.7768      & 0.8409       & 229.3613 \\
      3  & 26.89 & 23.869 & --     & --     & --     & --     & --     & --          & --          & --           & -- \\
      4  & 26.89 & 26.840 & 0.0966 & 0.0484 & 0.1066 & 0.0978 & 0.0135 & 278.3644    & 0.9174      & 0.9981       & 274.4376 \\
      5  & 18.21 & 18.185 & 0.1401 & 0.0708 & 0.1725 & 0.1275 & 0.0269 & 129.9897    & 0.7389      & 0.9986       & 142.6621 \\
      6  & 8.90  & 3.421  & 0.1390 & 0.1320 & 0.4181 & 0.0474 & 0.2188 & 64.0288     & 0.1134      & 0.3844       & 72.1730 \\
      7  & 0.61  & -1.025 & 0.0593 & 0.0533 & 0.1379 & 0.0239 & 0.0776 & 10.2867     & 0.1733      & -1.6803      & -42.8870 \\
      8  & 1.72  & 0.018  & 0.0896 & 0.0550 & 0.1730 & 0.0202 & 0.1023 & 19.2005 & 0.1166 & 0.0104  & 0.8928 \\
      9 & 10.62 & 3.142 & 0.1994 & 0.1406 & 0.7319 & 0.1154 & 0.3129 & 53.2518 & 0.1576 & 0.2959 & 27.2360 \\
      10  & 16.64 & 8.228  & 0.3363 & 1.0379 & 0.1522 & 0.0337 & 0.1155 & 49.4796     & 0.2214      & 0.4945       & 244.1543 \\
      11 & 42.14 & 32.229 & 0.4417 & 1.0604 & 0.2292 & 0.0935 & 0.2297 & 95.4041     & 0.4079      & 0.7648       & 344.6952 \\
      12 & 35.79 & 25.854 & 0.2384 & 0.1313 & 0.3531 & 0.1321 & 0.1748 & 150.1258    & 0.3741      & 0.7224       & 195.7154 \\
      13 & 32.97 & 24.630 & 0.5915 & 2.8489 & 0.1785 & 0.1024 & 0.1420 & 55.7396     & 0.5737      & 0.7470       & 240.5273 \\
      \bottomrule
      \end{tabular*}
  \end{subtable}
  \caption{Configurations, validation results, and distributional and efficiency metrics for varying proxy reference while fixed proxy expert.}
  \label{tab:rq2-1_full}
\end{table}

%% file: tables/rq2-2_3tables_version.tex
\begin{table}[htbp]
  \centering
  \scriptsize
  \setlength{\tabcolsep}{3.5pt}
  \begin{subtable}{\textwidth}
    \centering
    \caption{Model configurations and their identifiers.}
    \label{tab:rq2-2-configurations}
    \begin{tabular*}{\linewidth}{@{\extracolsep{\fill}}clll@{}}
        \toprule
        ID & Proxy expert & Proxy reference & Target \\
        \midrule
        1 & Qwen3-14B         & Qwen3-4B      & Qwen3-8B \\
        2 & Qwen3-8B          & Qwen3-4B & Qwen3-8B \\
        3 & Qwen3-4B-RL       & Qwen3-4B & Qwen3-8B \\
        4 & Qwen3-4B-SFT      & Qwen3-4B & Qwen3-8B \\
        5 & Qwen3-4B-Early-RL & Qwen3-4B & Qwen3-8B \\
        6 & Qwen3-4B-Late-RL  & Qwen3-4B & Qwen3-8B \\
        \bottomrule
      \end{tabular*}
  \end{subtable}

  \vspace{0.75em}

  \begin{subtable}{\textwidth}
    \centering
    \caption{Validation results for each configuration.}
    \label{tab:rq2-2-validation}
    \begin{tabular*}{\linewidth}{@{\extracolsep{\fill}}c*{9}{c}@{}}
        \toprule
        ID & AIME24 & AIME25 & AIME26 & SMT25 & CMIMC25 & HMMT25 (Feb) & HMMT25 (Nov) & HMMT26 (Feb) & Avg \\
        \midrule
        1 & 26.667 & 22.083 & 18.542 & 28.420 & 15.000 & 10.833 & 9.375  & 16.667 & 18.448 \\
        2 & 26.458 & 23.125 & 16.250 & 27.123 & 12.500 & 9.792  & 7.708  & 16.288 & 17.405 \\
        3 & 63.750 & 52.500 & 57.083 & 52.476 & 30.156 & 31.458 & 40.833 & 33.902 & 45.270 \\
        4 & 53.750 & 42.292 & 44.792 & 44.575 & 23.438 & 23.542 & 34.583 & 25.947 & 36.615 \\
        5 & 57.920 & 47.500 & 48.750 & 50.240 & 30.160 & 26.880 & 35.420 & 29.920 & 40.850 \\
        6 & 70.000 & 53.130 & 57.920 & 53.300 & 32.340 & 31.880 & 40.830 & 33.520 & 46.610 \\
        \bottomrule
      \end{tabular*}
  \end{subtable}

  \vspace{0.75em}

  \begin{subtable}{\textwidth}
    \centering
    \caption{Distributional and efficiency metrics for each configuration.}
    \label{tab:rq2-2-metrics}
    \begin{tabular*}{\linewidth}{@{\extracolsep{\fill}}c*{12}{c}@{}}
        \toprule
        ID & $G_P$ & $G_T$ & $K_P$ & $K_B$ & $K_I$ & $K_T$ & $K_{\mathrm{gap}}$ & $E_P$ & $R_{\mathrm{KL}}$ & $R_{\mathrm{perf}}$ & $E_T$ \\
        \midrule
        1 & 1.72  & 0.018  & 0.0896 & 0.0550 & 0.1730 & 0.0202 & 0.1023 & 19.2005 & 0.1166 & 0.0105 & 0.8928 \\
        2 & 0.61  & -1.025 & 0.0593 & 0.0533 & 0.1379 & 0.0239 & 0.0776 & 10.2867     & 0.1733      & -1.6803     & -42.887 \\
        3 & 26.89 & 26.840 & 0.0966 & 0.0484 & 0.1066 & 0.0978 & 0.0135 & 278.3644 & 0.9174      & 0.9981      & 274.4376 \\
        4 & 18.21 & 18.185 & 0.1401 & 0.0708 & 0.1725 & 0.1275 & 0.0269 & 129.9897    & 0.7391      & 0.9986      & 142.6621 \\
        5 & 22.09 & 22.420 & 0.0437 & 0.0465 & 0.0494 & 0.0452 & 0.0064 & 505.492     & 0.9158      & 1.0149      & 495.9848 \\
        6 & 32.28 & 28.180 & 0.1008 & 0.0719 & 0.1073 & 0.0853 & 0.0439 & 320.2381    & 0.7950      & 0.8730      & 330.3634 \\
        \bottomrule
      \end{tabular*}
  \end{subtable}
  \caption{Configurations, validation results, and distributional and efficiency metrics for varying proxy expert while fixed proxy reference.}
  \label{tab:rq2-2_full}
\end{table}

%% file: tables/rq3_3tables_version.tex
\begin{table}[htbp]
  \centering
  \scriptsize
  \renewcommand{\arraystretch}{0.88}
  \setlength{\tabcolsep}{3.5pt}
  \begin{subtable}{\textwidth}
    \centering
    \caption{Model configurations and their identifiers.}
    \label{tab:rq3-configurations}
        \begin{tabular*}{\linewidth}{@{\extracolsep{\fill}}clll@{}}
        \toprule
        ID & Proxy expert & Proxy reference & Target \\
        \midrule
        1  & Qwen3-1.7B-RL & Qwen3-1.7B & Qwen3-1.7B \\
        2  & Qwen3-1.7B-RL & Qwen3-1.7B & Qwen3-4B \\
        3  & Qwen3-1.7B-RL & Qwen3-1.7B & Qwen3-8B \\
        4  & Qwen3-1.7B-RL & Qwen3-1.7B & Qwen2.5-7B \\
        5  & Qwen3-1.7B-RL & Qwen3-1.7B & Qwen2.5-7B-Instruct \\
        6  & Qwen3-1.7B-RL & Qwen3-1.7B & Qwen2.5-7B-Math \\
        7  & Qwen3-4B-RL   & Qwen3-4B   & Qwen3-1.7B \\
        8  & Qwen3-4B-RL   & Qwen3-4B   & Qwen3-4B \\
        9  & Qwen3-4B-RL   & Qwen3-4B   & Qwen3-8B \\
        10 & Qwen3-4B-RL   & Qwen3-4B   & Qwen2.5-7B \\
        11 & Qwen3-4B-RL   & Qwen3-4B   & Qwen2.5-7B-Instruct \\
        12 & Qwen3-4B-RL   & Qwen3-4B   & Qwen2.5-7B-Math \\
        13 & Qwen3-4B-RL   & Qwen2.5-3B & Qwen3-1.7B \\
        14 & Qwen3-4B-RL   & Qwen2.5-3B & Qwen3-4B \\
        15 & Qwen3-4B-RL   & Qwen2.5-3B & Qwen3-8B \\
        16 & Qwen3-4B-RL   & Qwen2.5-3B & Qwen2.5-7B \\
        17 & Qwen3-4B-RL   & Qwen2.5-3B & Qwen2.5-7B-Instruct \\
        18 & Qwen3-4B-RL   & Qwen2.5-3B & Qwen2.5-7B-Math \\
        19 & Qwen3-4B-RL   & Qwen3-1.7B & Qwen3-1.7B \\
        20 & Qwen3-4B-RL   & Qwen3-1.7B & Qwen3-4B \\
        21 & Qwen3-4B-RL   & Qwen3-1.7B & Qwen3-8B \\
        22 & Qwen3-4B-RL   & Qwen3-1.7B & Qwen2.5-7B \\
        23 & Qwen3-4B-RL   & Qwen3-1.7B & Qwen2.5-7B-Instruct \\
        24 & Qwen3-4B-RL   & Qwen3-1.7B & Qwen2.5-7B-Math \\
        \bottomrule
    \end{tabular*}
  \end{subtable}

  \par
  \vspace{0.5em}

  \begin{subtable}{\textwidth}
    \centering
    \caption{Validation results for each configuration.}
    \label{tab:rq3-validation}
    \begin{tabular*}{\linewidth}{@{\extracolsep{\fill}}c*{9}{c}@{}}
        \toprule
        ID & AIME24 & AIME25 & AIME26 & SMT25 & CMIMC25 & HMMT25 (Feb) & HMMT25 (Nov) & HMMT26 (Feb) & Avg \\
        \midrule
        1  & 35.000 & 32.292 & 30.417 & 33.137 & 16.719 & 18.958 & 14.792 & 18.939 & 25.032 \\
        2  & 45.208 & 39.583 & 36.458 & 40.684 & 20.469 & 20.000 & 26.042 & 23.295 & 31.467 \\
        3  & 40.625 & 33.750 & 31.667 & 37.146 & 17.656 & 14.583 & 25.625 & 24.621 & 28.209 \\
        4  & 0.208  & 0.000  & 0.000  & 0.354  & 0.000  & 0.000  & 0.417  & 0.189  & 0.146 \\
        5  & 8.542  & 4.583  & 3.333  & 12.028 & 2.188  & 2.083  & 5.000  & 3.598  & 5.169 \\
        6  & 1.458  & 1.250  & 0.417  & 1.415  & 0.156  & 0.000  & 0.417  & 0.379  & 0.686 \\
        7  & 37.500 & 31.667 & 30.000 & 36.085 & 17.188 & 17.708 & 18.333 & 23.295 & 26.472 \\
        8  & 61.250 & 55.000 & 57.083 & 53.302 & 30.313 & 32.917 & 34.375 & 33.523 & 44.703 \\
        9  & 63.750 & 52.500 & 57.083 & 52.476 & 30.156 & 31.458 & 40.833 & 33.902 & 45.270 \\
        10 & 0.208  & 0.208  & 0.000  & 0.000  & 0.000  & 0.000  & 0.000  & 0.000  & 0.052 \\
        11 & 11.458 & 8.958  & 7.708  & 12.972 & 3.281  & 2.292  & 4.792  & 3.303  & 6.811 \\
        12 & 5.208  & 2.083  & 2.917  & 3.892  & 0.313  & 0.833  & 1.875  & 0.568  & 2.211 \\
        13 & 35.208 & 33.542 & 32.292 & 36.557 & 16.563 & 16.458 & 17.917 & 23.106 & 26.455 \\
        14 & 63.333 & 55.417 & 57.500 & 52.594 & 32.188 & 34.167 & 38.542 & 33.902 & 45.955 \\
        15 & 67.500 & 57.252 & 59.375 & 52.005 & 34.531 & 33.958 & 40.833 & 34.659 & 47.519 \\
        16 & 15.833 & 14.792 & 10.8333 & 14.623 & 2.813 & 3.958 & 1.250 & 4.167 & 8.534 \\
        17 & 15.416 & 11.875 & 11.875 & 17.453 & 3.594 & 3.125 & 5.208 & 4.356 & 9.113 \\
        18 & 9.583  & 10.000 & 12.292 & 11.439 & 3.281 & 2.292 & 1.250 & 2.083 & 6.527 \\
        19 & 41.458 & 34.375 & 31.042 & 35.142 & 17.969 & 20.208 & 18.333 & 22.159 & 27.586 \\
        20 & 63.333 & 54.583 & 56.250 & 53.774 & 30.000 & 35.625 & 39.792 & 32.197 & 45.694 \\
        21 & 62.917 & 49.792 & 56.250 & 51.061 & 30.938 & 31.458 & 40.417 & 31.439 & 44.284 \\
        22 & 0.000  & 0.000  & 0.000  & 0.000  & 0.000  & 0.000  & 0.000  & 0.000  & 0.000 \\
        23 & 6.458  & 5.208  & 3.958  & 9.316  & 2.188  & 1.250  & 3.750  & 3.030  & 4.395 \\
        24 & 8.542  & 3.333  & 6.250  & 4.953  & 1.094  & 0.833  & 0.833  & 2.083  & 3.490 \\
        \bottomrule
    \end{tabular*}
  \end{subtable}

  \par
  \vspace{0.5em}

  \begin{subtable}{\textwidth}
    \centering
    \caption{Distributional and efficiency metrics for each configuration.}
    \label{tab:rq3-metrics}
    \begin{tabular*}{\linewidth}{@{\extracolsep{\fill}}c*{9}{c}@{}}
        \toprule
        ID & $K_P$ & $K_B$ & $K_I$ & $K_T$ & $K_{\mathrm{gap}}$ & $E_P$ & $R_{\mathrm{KL}}$ & $R_{\mathrm{perf}}$ & $E_T$ \\
        \midrule
        1  & 0.0679 & 0.0000 & 0.0679 & 0.0673 & 0.0007 & 239.0280 & 0.9912 & 0.9927  & 239.4056 \\
        2  & 0.0631 & 0.1055 & 0.0766 & 0.0595 & 0.0148 & 257.2108 & 0.7768 & 0.8409  & 229.3613 \\
        3  & 0.0593 & 0.1110 & 0.0773 & 0.0587 & 0.0157 & 273.6931 & 0.7594 & 0.6025  & 166.5928 \\
        4  & 0.0644 & 0.4615 & 0.1573 & 0.0128 & 0.0898 & 252.0186 & 0.0814 & -0.1013 & -128.4375 \\
        5  & 0.0891 & 0.6562 & 0.1272 & 0.0581 & 0.0454 & 182.1549 & 0.4568 & -0.1042 & -29.1050 \\
        6  & 0.0457 & 0.6277 & 0.0966 & 0.0243 & 0.0449 & 355.1422 & 0.2516 & -0.0052 & -3.4568 \\
        7  & 0.1069 & 0.1233 & 0.1104 & 0.0935 & 0.0261 & 251.5435 & 0.8469 & 0.6527  & 187.7219 \\
        8  & 0.1003 & 0.0000 & 0.1003 & 0.1008 & 0.0009 & 268.0957 & 1.0050 & 0.9997  & 266.6964 \\
        9  & 0.0966 & 0.0484 & 0.1066 & 0.0978 & 0.0135 & 278.3644 & 0.9174 & 0.9981  & 274.4376 \\
        10 & 0.0868 & 0.3420 & 0.1137 & 0.0298 & 0.0742 & 309.7926 & 0.2621 & -0.0646 & -58.3221 \\
        11 & 0.1586 & 0.5935 & 0.1741 & 0.0962 & 0.0683 & 169.551 & 0.5523 & -0.0018 & -0.5095 \\
        12 & 0.0600 & 0.5497 & 0.0571 & 0.0369 & 0.0424 & 448.1667 & 0.6462 & 0.0536  & 39.0515 \\
        13 & 0.4528 & 1.0498 & 0.2185 & 0.0545 & 0.1865 & 96.4443  & 0.2494 & 0.4015  & 321.7431 \\
        14 & 0.4215 & 1.2987 & 0.1347 & 0.0728 & 0.0947 & 103.6062 & 0.5405 & 0.6443  & 386.4698 \\
        15 & 0.4191 & 1.2307 & 0.1592 & 0.0599 & 0.1265 & 104.1995 & 0.3763 & 0.6661  & 485.6260 \\
        16 & 0.4460 & 0.0991 & 0.4474 & 0.2111 & 0.2623 & 97.9148  & 0.4718 & 0.1544  & 31.9469 \\
        17 & 0.5731 & 0.1621 & 0.4755 & 0.2576 & 0.2353 & 76.1996  & 0.5417 & 0.0516  & 8.7461 \\
        18 & 0.5018 & 0.1513 & 0.5327 & 0.1436 & 0.3622 & 87.0267  & 0.2696 & 0.1318  & 40.0905 \\
        19 & 0.2623 & 0.0000 & 0.2623 & 0.1173 & 0.1436 & 136.4468 & 0.4472 & 0.5215  & 159.1304 \\
        20 & 0.2415 & 0.1238 & 0.3498 & 0.1519 & 0.1368 & 148.1988 & 0.4342 & 0.7788  & 183.5023 \\
        21 & 0.2384 & 0.1313 & 0.3531 & 0.1321 & 0.1748 & 150.1258 & 0.3741 & 0.7224  & 195.7154 \\
        22 & 0.2010 & 0.3742 & 0.6416 & 0.0397 & 0.3816 & 178.0597 & 0.0619 & -0.0500 & -45.0882 \\
        23 & 0.3275 & 0.5606 & 0.7529 & 0.1320 & 0.3568 & 109.284 & 0.1753 & -0.0689 & -18.678 \\
        24 & 0.2217 & 0.5617 & 1.0942 & 0.0356 & 0.4687 & 161.4344 & 0.0325 & 0.0760  & 76.4045 \\
        \bottomrule
      \end{tabular*}
  \end{subtable}
  \caption{Configurations, validation results, and distributional and efficiency metrics for varying target while fixed useful differences.}
  \label{tab:rq3_full}
\end{table}